\documentclass[lettersize,journal]{IEEEtran}
\usepackage{amsmath,amsfonts}
\usepackage{array}
\usepackage{textcomp}
\usepackage{stfloats}
\usepackage{url}
\usepackage{verbatim}

\usepackage{multirow}
\usepackage{amsmath}
\usepackage{amsthm}
\usepackage{amssymb}

\usepackage{amsfonts}
\usepackage{lscape}
\usepackage{rotating}      
\usepackage{arydshln}		
\usepackage{wrapfig}       

\usepackage[hidelinks,colorlinks=false,breaklinks=true,pagebackref=false,allcolors=black]{hyperref}
\usepackage{url}
\usepackage{setspace}
\usepackage{makecell}
\usepackage{stfloats}
\usepackage{booktabs}
\usepackage{multirow}
\usepackage{graphicx}

\usepackage{subfig}
\usepackage[numbers,sort&compress]{natbib}

\usepackage{xcolor}
\usepackage{algorithm}
\usepackage{algorithmicx}
\usepackage{algpseudocode}
\usepackage{amsmath}

\begin{document}

\title{Curricular Subgoals for Inverse Reinforcement Learning}

\author{Shunyu~Liu$^*$,
        Yunpeng~Qing$^*$,
        Shuqi Xu,
        Hongyan Wu,
        Jiangtao Zhang,
        Jingyuan Cong,
        Tianhao Chen,\\
        Yunfu Liu,
        Mingli~Song$^\dagger$
\thanks{$^*$These authors contributed equally.}
\thanks{$^\dagger$Corresponding author.}
\thanks{S.~Liu and M.~Song are with the Alibaba-Zhejiang University Joint Research Institute of Frontier Technologies, Zhejiang University, Hangzhou 310027, China (e-mail: liushunyu@zju.edu.cn, brooksong@zju.edu.cn).}
\thanks{Y.~Qing and J.~Cong are with the College of Computer Science and Technology, Zhejiang University, Hangzhou 310027, China (e-mail: qingyunpeng@zju.edu.cn, kcj51@zju.edu.cn).}
\thanks{H.~Wu, J.~Zhang, and T.~Chen are with the College of Software Technology, Zhejiang University, Hangzhou 310027, China (e-mail: wuhognyan@zju.edu.cn, zhjgtao@zju.edu.cn, tianhaochen@zju.edu.cn).}
\thanks{S.~Xu and Y.~Liu are with Alibaba Group, Hangzhou 310027, China (e-mail: xushuqi.xsq@alibaba-inc.com, yunfuliu@gmail.com).}
}

\markboth{\fontsize{5.3pt}{0pt}\selectfont{\quad This work has been submitted to the IEEE for possible publication. Copyright may be transferred without notice, after which this version may no longer be accessible.}}%
{Liu \MakeLowercase{\textit{et al.}}: Curricular Subgoals for Inverse Reinforcement Learning}


\maketitle

\begin{abstract}
Inverse Reinforcement Learning~(IRL) aims to reconstruct the reward function from expert demonstrations to facilitate policy learning, and has demonstrated its remarkable success in imitation learning. To promote expert-like behavior, existing IRL methods mainly focus on learning global reward functions to minimize the trajectory difference between the imitator and the expert. However, these global designs are still limited by the redundant noise and error propagation problems, leading to the unsuitable reward assignment and thus downgrading the agent capability in complex multi-stage tasks. In this paper, we propose a novel Curricular Subgoal-based Inverse Reinforcement Learning~(CSIRL) framework, that explicitly disentangles one task with several local subgoals to guide agent imitation. Specifically, CSIRL firstly introduces decision uncertainty of the trained agent over expert trajectories to dynamically select subgoals, which directly determines the exploration boundary of different task stages. To further acquire local reward functions for each stage, we customize a meta-imitation objective based on these curricular subgoals to train an intrinsic reward generator. Experiments on the D4RL and autonomous driving benchmarks demonstrate that the proposed methods yields results superior to the state-of-the-art counterparts, as well as better interpretability. Our code is available at \url{https://github.com/Plankson/CSIRL}.
\end{abstract}

\begin{IEEEkeywords}
Inverse reinforcement learning, curricular subgoals.
\end{IEEEkeywords}
\maketitle

\section{Introduction}

Inverse Reinforcement Learning~(IRL)~\cite{ng2000algorithms} has achieved impressive success in many decision-making tasks, such as video games~\cite{chan2021scalable}, robot manipulation~\cite{jabri2021robot}, and autonomous driving~\cite{zheng2021objective,aradi2020survey}.
Only provided with expert demonstrations without reward signals, the goal of IRL is to recover a reward function from observed behaviors for the desired optimal policy. 
One of the mainstream IRL approaches is adversarial IRL, which uses the generative adversarial network to obtain the expert-like policy that confuses the reward discriminator~\cite{mirza2014conditional,fu2017learning,yu2019multi}. Recently, some advanced approaches further develop IRL in a non-adversarial way, realizing both the reward and the policy learning in a single optimization function~\cite{reddy2019sqil,garg2021iq}. These previous works all attempt to infer the reward function to minimize the policy difference between the expert and the agent.

\begin{figure}[!t]
    \centering
    \includegraphics[width=0.49\textwidth]{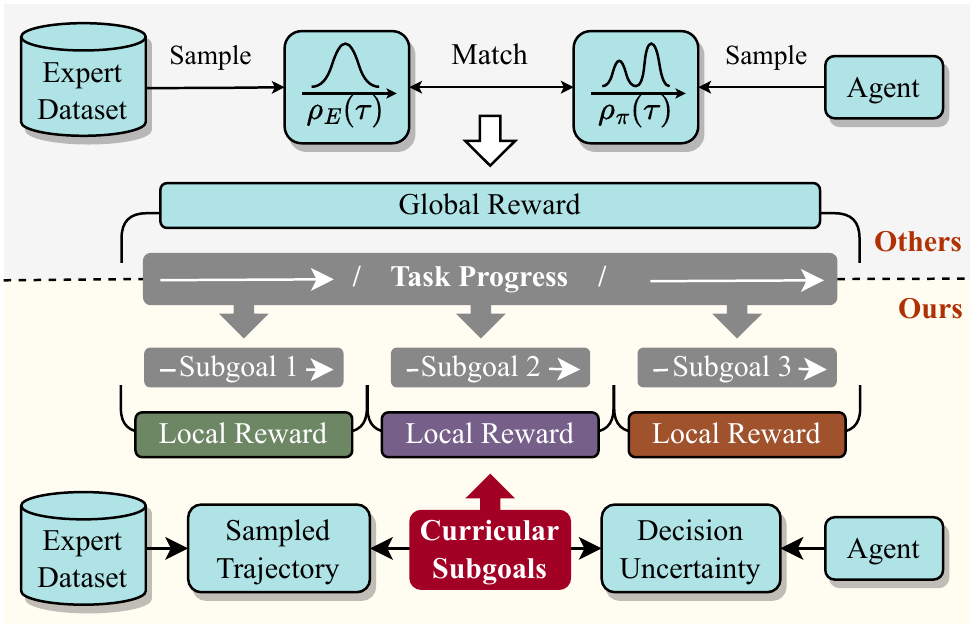}
    \caption{
    Comparing the proposed CSIRL method with existing methods. Existing methods construct a global reward function to match the policy distributions of the expert $\rho_E$ and the agent $\rho_\pi$ on the overall trajectory $\tau$, while CSIRL introduces curricular subgoals to disentangle one task with different local reward functions.
    }
    \label{fig:differ}
\end{figure}

Despite the encouraging results achieved, the reconstructed reward function of existing IRL methods inclines to evaluate the overall trajectory difference from a global perspective~\cite{vsovsic2018inverse,kostrikov2018discriminator,kostrikov2019imitation}, as depicted in the upper row of Figure~\ref{fig:differ}. For simple tasks, the directly learned global reward is enough to explain and reproduce the entire dataset of expert behavior. However, this global way is inflexible in complex tasks, which does not take into account the detailed learning difficulty along the task progress during training~\cite{levine2011nonlinear}. Especially in the multi-stage task with various environmental dynamics, exploration following the order of stages often introduces redundant noise from each other stage, which leads to an unsuitable assignment of the global reward and subsequently suboptimal results~\cite{shahryari2017inverse,lian2022anomaly}. For example, in a pick-and-place task of robot manipulation, the agent does not need to learn how to place unless it has learned how to pick. Moreover, policy matching based on the global reward function is inevitably limited by the error propagation problem~\cite{ross2010efficient,ross2011reduction,lindneractive}, where a slight discrepancy in the early stage of the task may lead to a severe error in the later stage. Thus, the matching difficulty between the overall trajectories of the agent and the expert will increase as the task progresses. As a result, the trained IRL agent often performs poorly in tasks that require long-term skills.

In this paper, we introduce a novel curricular subgoal-based IRL framework, termed as CSIRL, to explicitly decompose one complex task into several low-level subtasks. Unlike prior methods that directly realize policy matching in a global manner, the proposed CSIRL, as depicted in the lower row of Figure~\ref{fig:differ}, adopts a curriculum learning mechanism to guide agent imitation. The selected subgoals serve as a temporal task abstraction over expert trajectories, providing a series of simplified subtasks for agents to learn. In other words, CSIRL enables agents to explain expert trajectories in stages and further recover different local reward functions for each stage. These local reward functions are more compact than the single global one, effectively evaluating the agent behaviors based on its increasing capability. Therefore, the agent can use the assigned local rewards to progressively learn the final policy that matches expert demonstrations.

The proposed CSIRL comprises two key components, namely curricular subgoal selection and intrinsic reward generation. Technically, we adopt the decision uncertainty of the training agent to iteratively select the subgoals from expert trajectories. The curricular subgoals explicitly illustrate the exploration boundary during 
 different training stages. Then CSIRL further leverages these subgoals to reconstruct the local reward functions, consisting of the matching reward and intrinsic reward. The matching reward directly guides agent exploration towards the selected subgoals, while the intrinsic reward helps to explore the state space around expert demonstrations. We customize an additional meta-imitation learning objective to train such an intrinsic reward generator. The agent trained on this local reward function is then leveraged to determine the next subgoal.

Our main contribution is therefore a dedicated curricular subgoal-based IRL framework that enables multi-stage imitation based on expert demonstrations. Extensive experiments conducted on the D4RL and autonomous driving benchmarks show that our proposed CSIRL framework yields significantly superior performance to state-of-the-art competitors, as well as better interpretability in the training process. Moreover, the robustness analysis experiments show that CSIRL still maintains high performance even with only one expert trajectory.

The remainder of this manuscript is organized as follows. In Section~\ref{sec:relate}, we review some IRL topics highly related to this work. Then preliminaries of IRL are provided in Section~\ref{sec:background}. The proposed CSIRL is provided in Section~\ref{sec:method}. Experimental results, including the experiment settings, benchmark comparison and ablation study, are presented in Section~\ref{sec:result}. Finally, we conclude this manuscript with some possible future works in Section~\ref{sec:conclusion}.

\section{Related Works~\label{sec:relate}}

We briefly review here some recent advances that are most related to the proposed work in the context of inverse reinforcement learning and goal-based reinforcement learning. 

\subsection{Inverse Reinforcement Learning} 

Inverse reinforcement learning uses expert demonstrations to reconstruct the reward function for agent training, which has become a well-established paradigm for tackling Imitation Learning~(IL) problems~\cite{duan2017one,de2019causal,rajaraman2020toward,eraqi2022dynamic}. The IRL training can be formed as a min-max optimization on the reward and policy space, under which the expert obtains a high cumulative reward and the imitator learns an expert-like policy~\cite{ziebart2008maximum,ziebart2010modeling}.
\citet{ho2016generative} prove that IRL is the dual problem of occupancy measure matching. Moreover, they propose a solution based on a generative adversarial network. This adversarial training way is then widely studied by many other IRL approaches to reconstruct a more robust reward 
function~\cite{fu2017learning,yu2019multi,jeon2020regularized,orsini2021matters,bhattacharyya2022modeling} and improve sample efficiency~\cite{kostrikov2018discriminator,jarrett2020strictly}. 
Besides, many recent works also improve IRL in a non-adversarial way. \citet{reddy2019sqil} directly construct a simple binary reward and achieve the performance on par with the adversarial one. \citet{garg2021iq} avoid adversarial training by transforming the min-max optimization problem on the reward and policy into a single Q-function optimization, which obtains state-of-the-art results.

However, existing adversarial and non-adversarial IRL methods are still limited by learning long-term skills, especially in multi-stage tasks. The unbalance of exploration difficulty in different states causes improper reward assignments and suboptimal policy.
To tackle this problem, \citet{zhou2021inverse}  propose to learn language-based reward functions and corresponding optimal policies with input human language. It inspires the IRL community to leverage extra human supervision to improve IL training~\cite{mahmoudieh2022zero,sumers2021learning,zakka2022xirl}. On the contrary, our framework solves the problem without such supervision by autonomously detecting current exploration boundary during training, which can be viewed as the curricular subgoals to guide the training process.

\subsection{Goal-based Reinforcement Learning}
Goal-based reinforcement learning promotes agent learning by designing diverse explicit goals~\cite{kaelbling1993learning,andrychowicz2017hindsight,fang2019curriculum,5898415,7078931}. 
Current literature on goal-based RL achieves diverse goals by introducing the goal representation to the value function or policy ~\cite{schaul2015universal,pong2018temporal,wang2021reinforcement}.
Moreover, for tasks with goal-based sparse reward settings, current literature enhances the original reward with an intrinsic reward signal to implicitly encourage arriving at the goal state~\cite{andrychowicz2017hindsight,liu2022learn}. To arrange reasonable goals for current learning stages, \citet{chane2021goal}  construct a high-level policy based on the value function to predict intermediate states halfway, called imagined subgoals. \citet{nunez2022learning}  model the goal selection as an RL problem to utilize the Deep Q-Network framework~\cite{mnih2013playing} to select goals.
In the autonomous curriculum learning field~\cite{bengio2009curriculum}, adversarial methods are used to generate goals that are capable of being accessed after training~\cite{racaniere2019automated,florensa2018automatic}.

In the IRL field, several studies try to generate goals from expert trajectory to reconstruct reward function and obtain high expert-like behaviors. However, the subgoals they selected are all handcrafted and fixed in advance. \citet{ding2019goal} relabel the expert goal randomly from a fixed number of states after the current state. \citet{paul2019learning}  consider the equipartition of all goals along each trajectory and reconstruct the reward function according to the goals to perform DRL algorithms.
These frameworks ignore the varying learning difficulty along the task progress. This drawback makes it intractable for these approaches to be applied in more changeable environments.
\citet{vsovsic2018inverse}  propose an integrated Bayesian framework to predict the current subgoal of the agent, which is only workable for tasks with discrete action and state space. Our framework, practicable in continuous environments, autonomously generates the goals as curriculums according to the uncertainty of the trained agent on expert demonstration.

\section{Preliminaries~\label{sec:background}}

In this section, we formally define the IRL problem under the Markov decision process~(MDP). Then we introduce the Soft Actor-Critic~(SAC) algorithm.

\subsection{Markov Decision Process}

Markov Decision Process~(MDP) is an ideal abstracting to formulate reinforcement learning~(RL) tasks~\cite{puterman2014markov}, which can be described as a tuple $M = \langle\mathcal{S},\mathcal{A},P,r,\gamma,\rho_0\rangle$, where $\mathcal{S}$ represents the state space, $\mathcal{A}$ represents the action space, $P:\mathcal{S} \times \mathcal{A} \times \mathcal{S} \rightarrow \left[ 0,1\right]$ is the dynamics distribution, $r:\mathcal{S} \times \mathcal{A} \rightarrow \mathbb{R}$ is the reward function,  $\gamma \in (0 , 1)$ is the discount factor, and $\rho_0: \mathcal{S}\times\left[0,1\right]$ is the distribution of initial state $s_0$. 
At each timestep $t$, the agent observes the current state $s_t\in\mathcal{S}$ and selects an action $a_t\in\mathcal{A}$ drawn from the agent policy $\pi$, which causes a transition to the next state $s_{t+1}$ according to the dynamics distribution $P$. Moreover, the agent can gain a reward signal $r_t$. The goal of RL is to learn an optimal policy $\pi^*$ that maximizes the expected return:~$\pi^*=\arg\max_{\pi}\mathbb{E}_\pi\left[\sum_{k=0}^\infty\gamma^k r_{t+k}\right]$. 
 
Considering MDP from the perspective of IRL, the actual reward function $r$ is unavailable to the agent. Thus, IRL aims to recover the reward function from the given expert demonstrations for an expert-like optimal policy. This process can be abstracted as min-max optimization on the reward function space and state space. Specifically, the reward function is expected to maximize the cumulative reward to the expert policy, and the agent policy has to minimize the reward difference with the expert.

\subsection{Soft Actor-Critic}

Soft Actor-Critic~(SAC) is an off-policy algorithm for maximum entropy RL, which aims to maximize the expected reward while encouraging the agent to act as randomly as possible~\cite{haarnoja2018soft}. SAC comprises two key components: the critic network of soft Q-function $Q_\phi(s,a)$ with parameter $\phi$ and the actor network of policy $\pi_\theta(s)$ with parameter $\theta$. The critic network is used to estimate the expected return of the given state-action pair:
\begin{align}
\label{con::Q_def}
Q_\phi(s,a)=\mathbb{E}_{\pi}\left[\sum_{k=0}^\infty\gamma^k r(s_{t+k},a_{t+k}) \mid s_t=s,a_t=a\right].
\end{align}
We use the soft Bellman residual to train the critic network:
\begin{align}
    \label{con::Q_optim}
\mathcal{L}_Q(\phi)=\mathbb{E}_{\mathcal{B}}\left[\left(Q_\phi(s,a)-\left(r(s,a)+\gamma V_{\bar{\phi}}(s')\right)\right)^2\right],
\end{align}
where $\mathcal{B}$ denotes the replay buffer of the transitions and the value function is calculated by \begin{align}
V_{\hat{\phi}}(s')=\mathbb{E}_{a'\sim\pi}\left[Q_{\hat{\phi}}(s',a')-\alpha\log(\pi(a'|s')\right],
\end{align}
where $\alpha$ is the temperature coefficient and $\hat{\phi}$ is the parameter of the target critic network. The policy network is updated towards soft Q-function as follows, which guarantees to achieve an improved policy~\cite{haarnoja2018soft}:
\begin{align}
\label{con::pi_optim}
    \mathcal{L}_\pi(\theta,\phi)=\mathbb{E}_{s\sim \mathcal{B},a\sim\pi_\theta}\left[\alpha \log \pi_\theta(a|s)-Q_\phi(s,a)\right].
\end{align}
Finally, SAC trains the agent by iteratively updating the critic network and the actor network. Our CSIRL implementation uses SAC as a basic backbone for its robust performance.

\begin{figure*}[!t]
    \centering
    \includegraphics[width=1.0\textwidth]{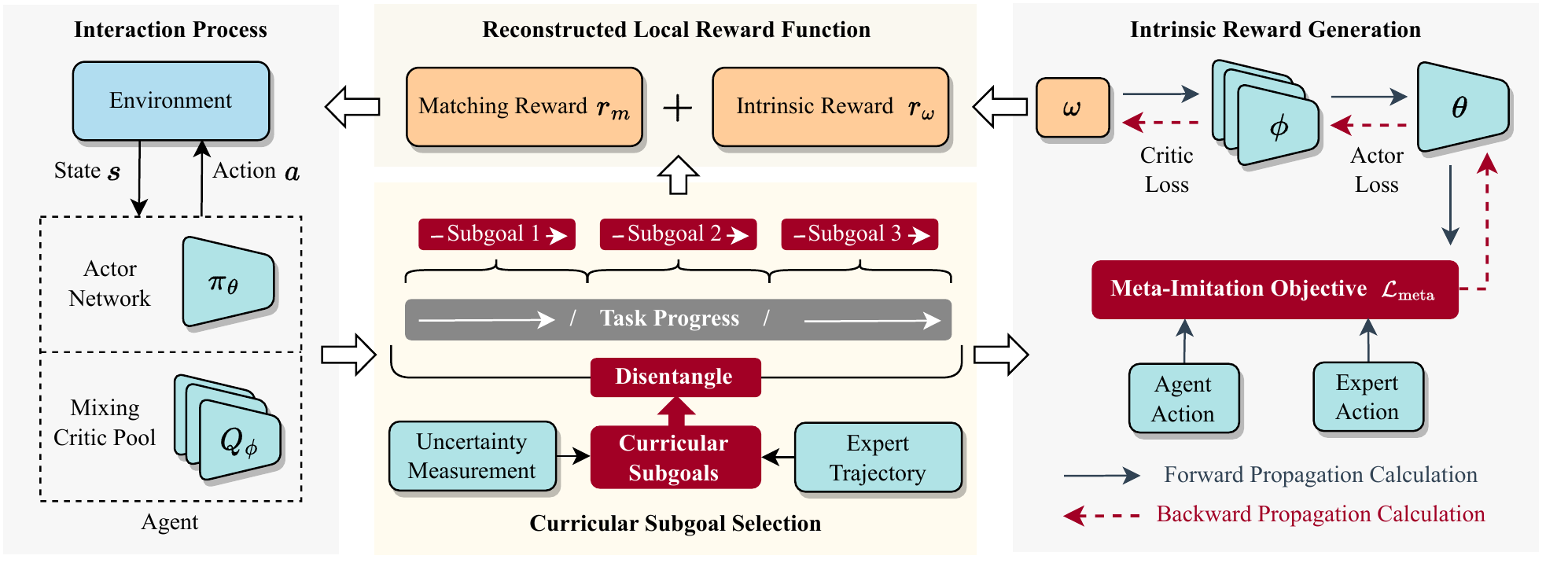}
    \caption{An illustrative diagram of the proposed Curricular Subgoal-based Inverse Reinforcement Learning~(CSIRL) method.}
    \label{fig::subgoal}
\end{figure*}

\section{Method~\label{sec:method}}

In what follows, we first provide an overview of the training process based on the proposed curricular subgoal-based inverse reinforcement learning~(CSIRL) framework. Then we further detail the CSIRL framework,  as shown in Figure~\ref{fig::subgoal}.

\subsection{Overview}
Our framework consists of two key components: curricular subgoal selection and intrinsic reward generation.
Curricular subgoal selection is based on agent uncertainty over expert demonstrations, where the uncertainty is measured by the variance of several critic networks. Then we can reconstruct the local matching reward towards the selected subgoals.
In the intrinsic reward generation step, a local intrinsic reward generator is designed to fine-tune the local reward function. The generator is trained over a meta-imitation learning objective to help exploration around expert demonstrations.
This combined local reward is then used for SAC training.
With these two components, CSIRL obtains a policy achieving expert-like performance.

\subsection{Curricular Subgoal Selection}
\label{sec::cs}
The curricular subgoals are selected based on the decision uncertainty, which is then utilized to reconstruct the local reward function. 
The curricular subgoal $g\in\mathcal{S}$ is defined as the state that appears in the expert dataset $\mathcal{D}=\left\{\tau^e|\tau^e=\left\{(s_t^e,a_t^e,s_{t+1}^e)_{t=0}^T\right\}\right\}$ with the trajectory length of $T$. 
This definition comes from the fact that the agent imitation can be viewed as accomplishing the task by advancing along the expert behaviors in a complete expert trajectory. During this process, there exist some puzzling states for the current imitator to reach. These states, called curricular subgoals, can inversely reconstruct the reward signal and further encourage the agent exploration towards these subgoals.

One key issue to be considered is the strategy for selecting reasonable curricular subgoals that are neither too hard nor too simple. To address this problem, we propose a method based on the uncertainty of the mixing critic pool to identify goals with appropriate difficulty. 
Intuitively, the more significant uncertainty on a given state $s$, the less information the agent has gained about the state space near $s$.
At each time during training, the uncertainty of the agent along the task progress gradually increases from the early stages to the later stages. This phenomenon occurs because of the discrepancy in sampling difficulty at different stages. Therefore, if the uncertainty of expert behavior and the subsequent ones exceed a fixed threshold, this behavior is probably at the current exploration boundary of the agent. Q-function that we have discussed in Section~\ref{sec:background} maintains the expected return value of the input state-action pairs, which can be utilized to evaluate such uncertainty. Specifically, a mixing critic pool can address the uncertainty-measuring problem. The mixing critic pool under the subgoal $g$ can be represented as:
\begin{align}
\mathbf{Q}_\phi(s,a):=\left[Q_{\phi_1}(s,a),Q_{\phi_2}(s,a),\cdots,Q_{\phi_n}(s,a)\right],
\end{align}
where $Q_{\phi_i}$ is the $i$-th critic network approximating the Q-function with parameter $\phi_i$ and $n$ is the number of independent critic networks. The critic networks in the pool are updated independently according to SAC during training. 
That is, with the transitions $(s,a,s')$ sampled from the replay buffer $\mathcal{B}$, each critic network $Q_{\phi_i}$ can be optimized using the soft Bellman residual $\mathcal{L}_Q(\phi_i)$ according to Equation~(\ref{con::Q_optim}).  The mixing critic pool tracks the performance of imitator policy. Moreover, the diversity of the critic outputs provides an approximation of the uncertainty.

Specifically, let $s$ be the state that we need to measure the uncertainty of the agent on. The agent chooses action $a$ drawn from policy $\pi_\theta(\cdot|s)$ with parameter $\theta$. The uncertainty $U:\mathcal{S}\rightarrow \mathbb{R}$ can be computed as follows:
\begin{align}
\label{con::un}
    U(s)=\mu\left(\mathbf{Q}_\phi(s,a)\right),
\end{align}
where $\mu$ is the variance operation over the outputs of different critic networks.
 
To choose curricular subgoals with appropriate difficulty, an expert trajectory $\tau^e=\left\{(s^e_t,a^e_t,s^e_{t+1})\right\}_{t=0}^T$ is assigned to reveal and further disentangle the task. At the start of each training round, the agent has learned the policy $\pi_\theta$ and the critic network $\mathbf{Q}_\phi$ about reaching the previous subgoal $g$.
Feeding each expert state to $\pi_\theta$ and $\mathbf{Q}_\phi$, the uncertainty of trajectory $\mathbf{U}(\tau^e)$ is obtained:
\begin{align}
    \mathbf{U}(\tau^e):=\left[U(s^e_0),U(s^e_1), \cdots, U(s^e_T)\right].
\end{align}
Then the $\mathbf{U}(\tau^e)$ is scanned from back to end to seek the state with an immediately high uncertainty value. To decrease approximation error for stable evaluation, a soft average value $U_p^{\textrm{avg}}$ at position $p$ is introduced, which is set to $U(s^e_0)$ initially and can be updated while scanning to $s^e_p$ by 
\begin{align}
    U_p^{\textrm{avg}}\leftarrow \sigma U_{p-1}^{\textrm{avg}}+(1-\sigma)U(s^e_p),
\end{align}
where $\sigma$ is a soft-updating coefficient. During scanning, the first state $s^e_p$ that satisfies $U(s^e_p) / U_{p-1}^{\textrm{avg}}>\delta$ is detected as the subgoal $g$ for the next round of training  where $\delta$ is a uncertainty threshold. This strategy ensures that the following subgoal $g$ is the first state with an immediately high uncertainty value after previous subgoals. The expert demonstrations from $s^e_0$ to $g$ form a disentangled curriculum progress containing a series of stages. This progress is a feasible curriculum with appropriate difficulty for the current agent. 
If there is no state uncertainty satisfying this condition, the training process can be stopped since the uncertainty of all elements of $\mathbf{U}(\tau^e)$ have decreased to an acceptable range.

As for local reward reconstruction, the curricular subgoals provide an intuitive approach of reconstituting the local matching reward function using potential-based shaping~\cite{ng1999policy} as
\begin{align}
\label{con::match_reward}
    r_m(s,a,s',g)=\psi(s',g)-\gamma\psi(s,g),
\end{align}
where the agent takes action $a$ at state $s$ and reaches state $s'$, $\gamma$ is the discount factor and $\psi$ is a state-evaluating function that can be represented as
\begin{align}
    \label{con::repre}
    \psi(s,g)=\parallel\!h(s)\!-\!h(g)\!\parallel_2^2,
\end{align}
where $h$ is a representation function of state. Let $p$ represents the closest state from $s$ in expert trajectory. Equation~(\ref{con::repre}) ensures that if $p$ appears before $g$, then the agent is encouraged to act towards subgoal $g$ by $\parallel h(s)-h(g) \parallel_2^2$. In the case that $p$ occur in the follow-up states of $g$, the agent is expected to move directly towards the final state $s^e_T$ by $\psi(s)=\parallel h(s)-h(s^e_T) \parallel_2^2$. This matching reward function guides the agent to advance toward the subgoal and further accomplish the whole task. Based on this local reward function, we utilize the SAC algorithm to train the imitator and obtain the mixing critic pool and policy necessary for the next round of subgoal selection and training.

\subsection{Intrinsic Reward Generation}
An intrinsic reward generator $r_\omega(s,a,g)$ parameterized with $\omega$ is proposed to assist agent training. The matching reward $r_m$ directly measures the similarity of the current state and subgoal in the long horizon. To further improve the exploration efficiency and imitation accuracy of the agent, the intrinsic reward models expert preference and further fine-tune the local reward model. The optimization of $r_\omega$ captures the implicit gradients and dependencies between the reconstructed reward and agent policy from policy evaluation and improvement of SAC. The intrinsic reward explicitly encourages the agent to act like an expert from the reward perspective with the meta-imitation objective.
More specifically, this meta-imitation objective with BC form is proposed as:
\begin{align}
\label{con::meta}
    \mathcal{L}_{\text{meta}}(\omega,\theta)=\mathbb{E}_{(s,a)\in \mathcal{D},\hat{a}\sim \pi_\theta(\cdot|s)} \frac{1}{2}\left( \hat{a}-a\right)^2,
\end{align}
where $\pi_\theta$ is the imitator policy. 
However, there is no term that directly contains intrinsic reward $r_\omega$. To tackle this issue, we analyze the gradient of reward function according to the chain rule in RL training. We further give a detailed derivation of how the simple meta-imitation objective in the RL training stage can optimize the intrinsic reward. 

To construct the connection between the intrinsic reward generator parameter $\omega$ and the meta-imitation objective in SAC backbone training, we first start from the mixing critic pool $\mathbf{Q}_\phi$. As in policy evaluation process, each critic network entity $Q_{\phi_i}$ with parameter $\phi_i$ is optimized by the soft Bellman residual loss $\mathcal L_Q(\phi_i,\omega)$ formed in Equation~(\ref{con::Q_optim}). Notice that the combined local reward $r_m(s,a,s',g)+r_\omega(s,a,g)$ replaces the original $r(s,a)$ in Equation~(\ref{con::Q_optim}). Therefore, the critic is updated by $\phi' = \phi-\alpha_1\nabla_{\phi} \mathcal L_Q(\phi,\omega)$ where $\phi'$ is the parameter of the updated critic entity and $\alpha_1$ is the learning rate of critic network. The critic network entity optimization naturally defines the partial derivative of $\phi'$ over $\omega$:
\begin{align}
\nabla_\omega \phi'=\nabla_\omega\left(\phi-\alpha_1\nabla_\phi\mathcal{L}_Q(\phi,\omega)\right)=-\alpha_1\nabla_\omega\nabla_\phi\mathcal{L}_Q(\phi,\omega)
\label{con::r_q}
\end{align}
This equation illustrates how the gradient can be backpropagated from critic network $\phi$ to intrinsic reward generator $\omega$. 

Moreover, the policy improvement step implicitly constructs the partial derivative of the updated actor $\theta'$ over the updated critic $\phi'$.
The policy $\pi_\theta$ is updated utilizing the updated critic $Q_{\phi'}$ via $\mathcal L_\pi(\theta,\phi')$ defined in Equation~(\ref{con::pi_optim}). Similar to Equation~(\ref{con::r_q}), the dependence between the updated actor $\theta'$ and the updated critic $\phi'$ can be captured as:
\begin{align}
\label{con::q_pi}
    \nabla_{\phi'}\theta'=\nabla_{\phi'} \left(\theta-\alpha_2\nabla_{\theta}\mathcal\mathcal{L}_\pi(\theta,\phi')\right)=-\alpha_2\nabla_{\phi'}\nabla_{\theta}\mathcal{L}_\pi(\theta,\phi'),
\end{align}
where $\alpha_2$ is the learning rate of actor.

Finally, the optimized actor $\pi_{\theta'}$ is leveraged to compute the BC-style meta-imitation loss function $\mathcal L_\text{meta} (\omega, \theta')$. Since the previous policy evaluation and improvement steps have already established the explicit chain rule components as described in Equation~(\ref{con::r_q}) and~(\ref{con::q_pi}), the intrinsic reward generator parameter $\omega$ can be optimized via
\begin{align}
    \omega'&=\omega-\alpha_3\nabla_\omega\mathcal{L}_\text{meta}(\omega,\theta')\\
    &=\omega-\alpha_3\nabla_{\theta'}\mathcal{L}_\text{meta}(\omega,\theta')\nabla_{\phi'}\theta'\nabla_\omega \phi',
\end{align}
where $\omega'$ is the updated intrinsic reward model and $\alpha_3$ is the learning rate of the intrinsic reward. 
Namely, the intrinsic reward is learning from RL training. This meta-learning optimization captures the hidden mechanism of how the local reward influences agent learning. Moreover, the simple meta-imitation target $\mathcal L_\text{meta}$ guides the local intrinsic reward optimization. The intrinsic reward models expert preference and further induces the expert-like policy.
We summarize the entire training procedure in Algorithm~\ref{alg:alg}.

\algdef{SE}[SUBALG]{Indent}{EndIndent}{}{\algorithmicend\ }%
\algtext*{Indent}
\algtext*{EndIndent}
\algnewcommand{\LineComment}[1]{\State \textcolor{gray}{\(\triangleright\) #1}}

\begin{algorithm}[!t]
    \caption{CSIRL}
    \label{alg:alg}
    \begin{flushleft}
        \textbf{Initialize:} Actor Network $\pi_{\theta}$, Mixing Critic Pool $\mathbf{Q}_\phi$, Exp- ert Dataset $\mathcal{D}$, Replay Buffer $\mathcal{B}$, Uncertainty Threshold $\delta$. \\
    \end{flushleft}
    \begin{algorithmic}[1]
    \Repeat
    \LineComment{Curricular Subgoal Selection}
    \State Sample expert trajectory $\tau^e \sim \mathcal{D}$ 
    \State Compute decision uncertainty $\mathbf{U}(\tau^e)$ and $U_p^{\textrm{avg}}$
    \State Select current subgoal
    $$g=\mathop{\arg\min}_{s^e_p\in\tau^e} \left\{p \mid  \frac{U(s^e_p)}{U^{\textrm{avg}}_{p-1}}>\delta \right\}$$
    \State Reconstruct matching reward $r_m$
    \LineComment{Intrinsic Reward Generation}
    \State Update generator $\omega \leftarrow\omega-\alpha_3\nabla_\omega\mathcal{L}_\text{meta}(\omega,\theta)$
    \State Reconstruct intrinsic reward $r_\omega$
    \LineComment{SAC Training Process}
    \State Sample transition $(s,a,s')$ with policy $\pi_\theta(\cdot|s)$
    \State Reconstruct local reward  $r = r_m + r_\omega$
    \State $\mathcal{B}\leftarrow\mathcal{B}\cup\left\{(s,a,s', r)\right\}$
    \State $\phi_i \leftarrow\phi_i-\alpha_1\nabla_{\phi_i} \mathcal{L}_Q(\phi_i,\omega)$ for $i\in\left\{1,2,\cdots,n\right\}$
    \State $\theta \leftarrow\theta-\alpha_2\nabla_\theta \mathcal{L}_\pi(\theta,\phi)$
    \Until converge
\end{algorithmic}
\end{algorithm}

\section{Experiments~\label{sec:result}}

In this section, we first detail the hyperparameter in CSIRL, the compared baselines and the environments for experiments. Then we report the comparison results and ablation studies, and further present the visualization analysis.

\subsection{Hyperparameter Settings}
Our proposed CSIRL starts training after 1280 interactions with random policy to initialize the experimental replay buffer.
Batches of 64 agent state-action pairs sampled from the replay buffer with the size of 1M and batches of 32 expert state-action pairs sampled from the expert dataset are required in each training iteration. The mixing critic pool contains $N=5$ network entities. The structures of entities of the critic networks, actor network, and intrinsic reward network are all full-connected neural networks with a layer number of 2 and a hidden dimension of 256. The critic and actor network are all updated with Adam optimizer with a learning rate of $3\times 10^{-4}$. We set a target critic network for each critic network and soft-update the target network with a soft-update rate of $0.05$.
The intrinsic reward network meta-learning update is implemented with 
a fixed critic and actor to calculate the gradient needed for the chain rule in each iteration. The fixed critic and actor perform manual gradient descent with a learning rate of $0.001$ while the intrinsic reward network is updated using Adam optimizer with a learning rate of $0.01$.
For curricular subgoal selection, we set the threshold $\delta=5.0$ to detect the immediately high uncertainty value. The soft update rate $\sigma$ for the soft average uncertainty value $U^p_{\textrm{avg}}$ is set to $0.2$.

\subsection{Baselines}
Our methods are compared with various state-of-the-art methods: (1)~the vanilla behavioral cloning~(BC)~\cite{pomerleau1991efficient}, (2)~the GAIL method~\cite{ho2016generative}, which utilizes GAN to construct expert behavior generator and discriminator, (3)~the DAC method~\cite{kostrikov2018discriminator}, which implies off-policy RL method to remove GAN-based reward bias, (4)~the ValueDICE method~\cite{kostrikov2019imitation}, which learns the imitator policy with a complete off-policy objective from the original distribution ratio estimation, (5)~the SQIL method~\cite{reddy2019sqil}, which manually assigns a positive reward signal only on expert state-action pair and then performs Q-Learning, and (6)~the IQ-Learn method~\cite{garg2021iq} which directly learning the Q-function from the expert demonstration.

\subsection{Scenarios and Datasets}
To demonstrate the effectiveness of the proposed CSIRL framework, we conduct experiments on the D4RL benchmark~\cite{fu2020d4rl} and the autonomous driving benchmark~\cite{highway_env}. 

(1) \textbf{D4RL Benchmark}.
We evaluate CSIRL on two D4RL environments, including the Mujoco task of \emph{AntUMaze} and the Androit task of \emph{Relocate}. 

\begin{figure}[!t]    
    \centering
    \subfloat[AntUMaze]{\includegraphics[width=0.23\textwidth]{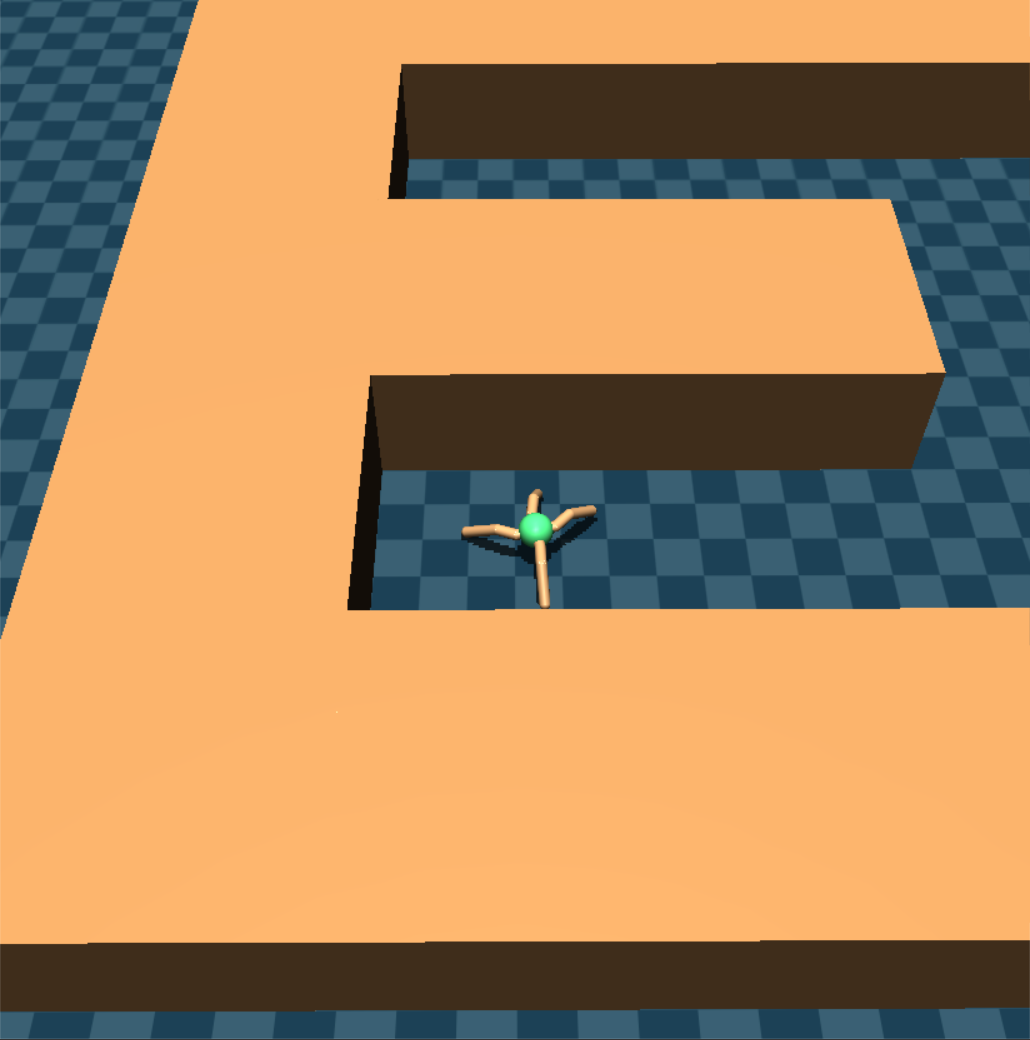}\label{fig::antumaze}}
    \hspace{0.02\linewidth}
    \subfloat[Relocate]{\includegraphics[width=0.23\textwidth]{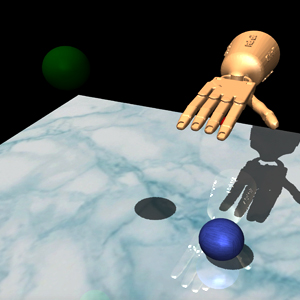}\label{fig::relocate}}
    \caption{
    Illustration of the D4RL benchmark.
    }
    \label{fig::d4rl}
\end{figure}

\begin{itemize}
    \item \textbf{AntUMaze}. In the \emph{AntUMaze} scenario, as shown in Figure~\ref{fig::antumaze}, the agent manipulates an ant robot through a U-shaped maze. The state space contains the location information of the ant robot and its specific joint information, $s\in\mathbb{R}^{29}$. The action $a\in\mathbb{R}^8$ controls the angle and speed of the ant robot joints.
    \item \textbf{Relocate}. In the \emph{Relocate} scenario, as shown in Figure~\ref{fig::relocate}, the agent is a hand robot that has to move a ball to a target position~\cite{Rajeswaran-RSS-18}. It should reach the ball position, grab it, and then try to grab it to the desired place. The state $s\in\mathbb{R}^{39}$ contains the hand joint information and the ball position. And the action $a\in\mathbb{R}^{30}$ control the angle and speed of the hand robot joints.
\end{itemize}

(2) \textbf{Autonomous Driving Benchmark}.
The autonomous driving environment is partially observable where the agent can only observe its own and several closest vehicles. The internal driving strategies~(changing speed and lane) of other vehicles are uncertain and unavailable to the agent. The agent must learn a diverse policy to reach the final destination within the specified time without collisions.
Different autonomous driving scenarios have the same structure of state space and the same action space. The state space is constructed by the kinematic information of ego vehicle and the $m$ closest vehicles: $\textbf{s}=\left[\textbf{v}_0,\textbf{v}_1,\cdots,\textbf{v}_m\right]$, where $\textbf{v}_0$ represents the ego vehicle information and $\textbf{v}_1,\cdots, \textbf{v}_m$ represents other $m$ closest cars. For different task, $\textbf{v}_i$ may contain different vehicle features. $\textbf{v}_i$ can include presence information $p$, vehicle position $x,y$, speed information $v_x,v_y$ and vehicle head angle $\sin\theta,\cos\theta$. If $\textbf{v}_i$ contains $n$ elements, then $s\in\mathbb{R}^{n(m+1)}$. Meanwhile, 
different from discrete action settings in which the action space only contains several elements that correspond to a series of continuous actions, 
in continuous tasks, the agent must directly control the acceleration and the steering angle of vehicles. Therefore, the action can be represented as $a\in \mathbb{R}^2$. Here we give a detailed description of the goal of each scenario and the specific state space structure.

\begin{figure}[!t]
    \centering
    \subfloat[Highway-fast]{\includegraphics[width=0.23\textwidth]{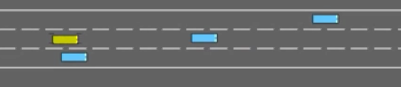}\label{fig::h}}
    \hspace{0.02\linewidth}
    \subfloat[Merge]{\includegraphics[width=0.23\textwidth]{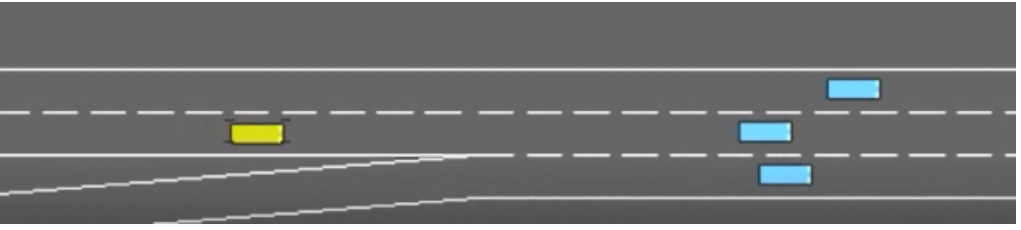}\label{fig::m}}\\
    \subfloat[Roundabout]{\includegraphics[width=0.23\textwidth]{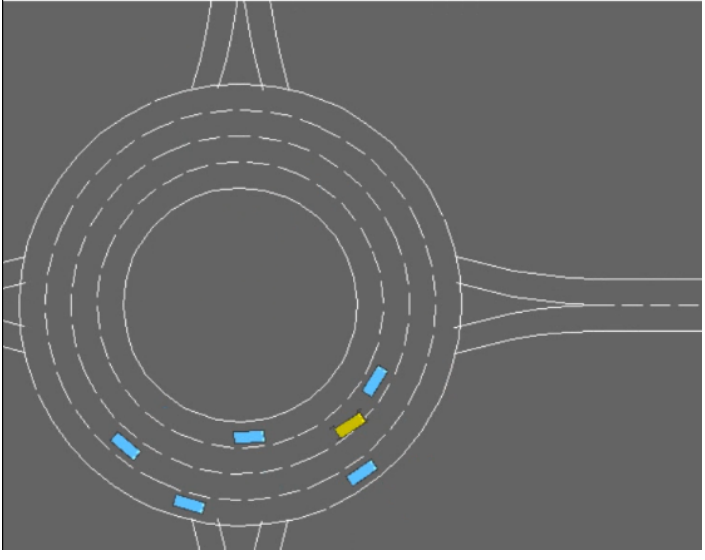}\label{fig::r}}
    \hspace{0.02\linewidth}
    \subfloat[Intersection]{\includegraphics[width=0.23\textwidth]{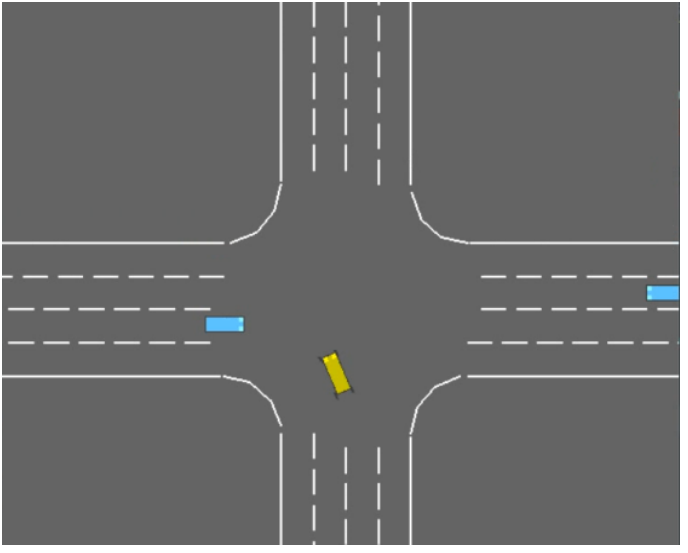}\label{fig::i}}
    \caption{
    Illustration of the Autonomous driving benchmark.
    }
    \label{fig::highway}
\end{figure}
\begin{itemize}
    \item \textbf{Highway-fast}. In the \emph{highway-fast} scenario, a controlled vehicle is required to drive on the highway, as shown in Figure~\ref{fig::h}. Specifically, this task simulates the actual highway scene in that the controlled vehicle has to keep going straight at high speed on the through lane while avoiding collision with other vehicles by changing lanes. The state of this task contains the ego and other $m=5$ closest vehicles: $\textbf{s}=\operatorname{costncat}(\left[\textbf{v}_0,\textbf{v}_1\cdots, \textbf{v}_5\right])\in\mathbb{R}^{30}$, where $\textbf{v}=\left[p,x,y,v_x,v_y\right]$.
    \item \textbf{Merge}. As illustrated in Figure~\ref{fig::m}, this task \emph{merge} simulates the intersection of the main lane and a merging lane. The autonomous vehicle must continue driving in the road while avoiding colliding with other vehicles on both the main and merging lanes. The state of this task contains the ego and other $m=5$ closest vehicles: $\textbf{s}=\operatorname{concat}(\left[\textbf{v}_0,\textbf{v}_1\cdots, \textbf{v}_5\right])\in\mathbb{R}^{30}$, where $\textbf{v}=\left[p,x,y,v_x,v_y\right]$.
    \item \textbf{Roundabout}.
    The \emph{roundabout} scenario is much more challenging than the previous two introduced tasks. As shown in Figure~\ref{fig::r}, in this scenario, the controlled vehicle has to drive smoothly into the roundabout, follow the bend until the second exit, and drive out of the roundabout from this exit, during which the agent also needs to avoid the collision. It should be noted that to simulate the actual environment, we increase the number of lanes in the roundabout to $4$ and increase the vehicle count in the roundabout, which makes the task much different from the original one. The state of this task contains the ego and other $m=5$ closest vehicles: $\textbf{s}=\operatorname{concat}(\left[\textbf{v}_0,\textbf{v}_1\cdots, \textbf{v}_5\right])\in\mathbb{R}^{42}$, where $\textbf{v}=\left[p,x,y,v_x,v_y,\cos\theta,\sin\theta\right]$.
    \item \textbf{Intersection}.
    In this scenario, the controlled vehicle has to turn at the intersection and drive into the correct lane while avoiding collision with vehicles in all directions, as shown in Figure~\ref{fig::i}.
    It should be noted that to simulate the actual environment we increase the one-way lane on each main road to $2$.
    The state of this task contains the ego and other $m=5$ closest vehicles: $\textbf{s}=\operatorname{concat}(\left[\textbf{v}_0,\textbf{v}_1\cdots, \textbf{v}_5\right])\in\mathbb{R}^{42}$, where $\textbf{v}=\left[p,x,y,v_x,v_y,\cos\theta,\sin\theta\right]$.
\end{itemize}

For D4RL benchmark, we directly leverage the expert dataset provided by D4RL~\cite{fu2020d4rl}.
For autonomous driving benchmark, we generate expert demonstrations by running an optimal planning algorithm~\cite{hren2008optimistic}.  
Then we sample datasets of varying trajectory numbers from the expert demonstrations to test the compared baselines. 
The expert dataset with more trajectories involves more situations in real interaction, which implicitly represents the expert policy to the greatest extent.

\begin{figure}[!t]    
    \centering
\subfloat{\quad\includegraphics[scale=0.35]{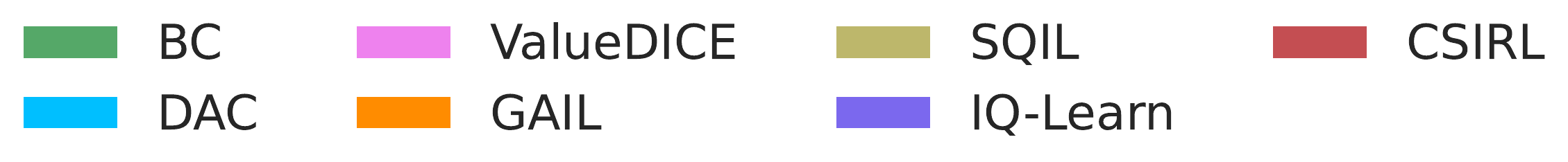}}\\
    \addtocounter{subfigure}{-1}
    \subfloat[AntUMaze]{\includegraphics[width=0.235\textwidth]{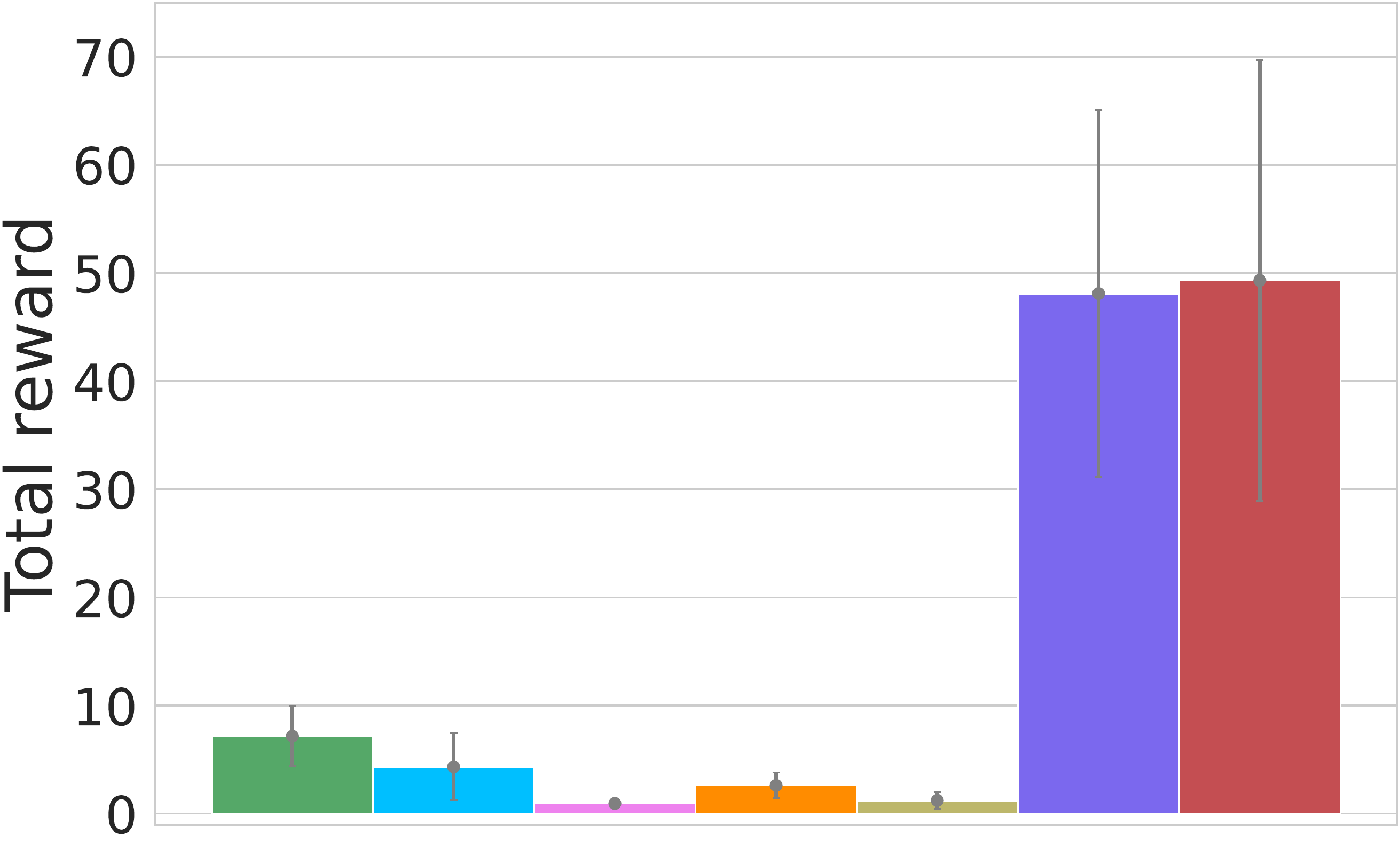}}
    \hspace{0.01\linewidth}
    \subfloat[Relocate]{\includegraphics[width=0.235\textwidth]{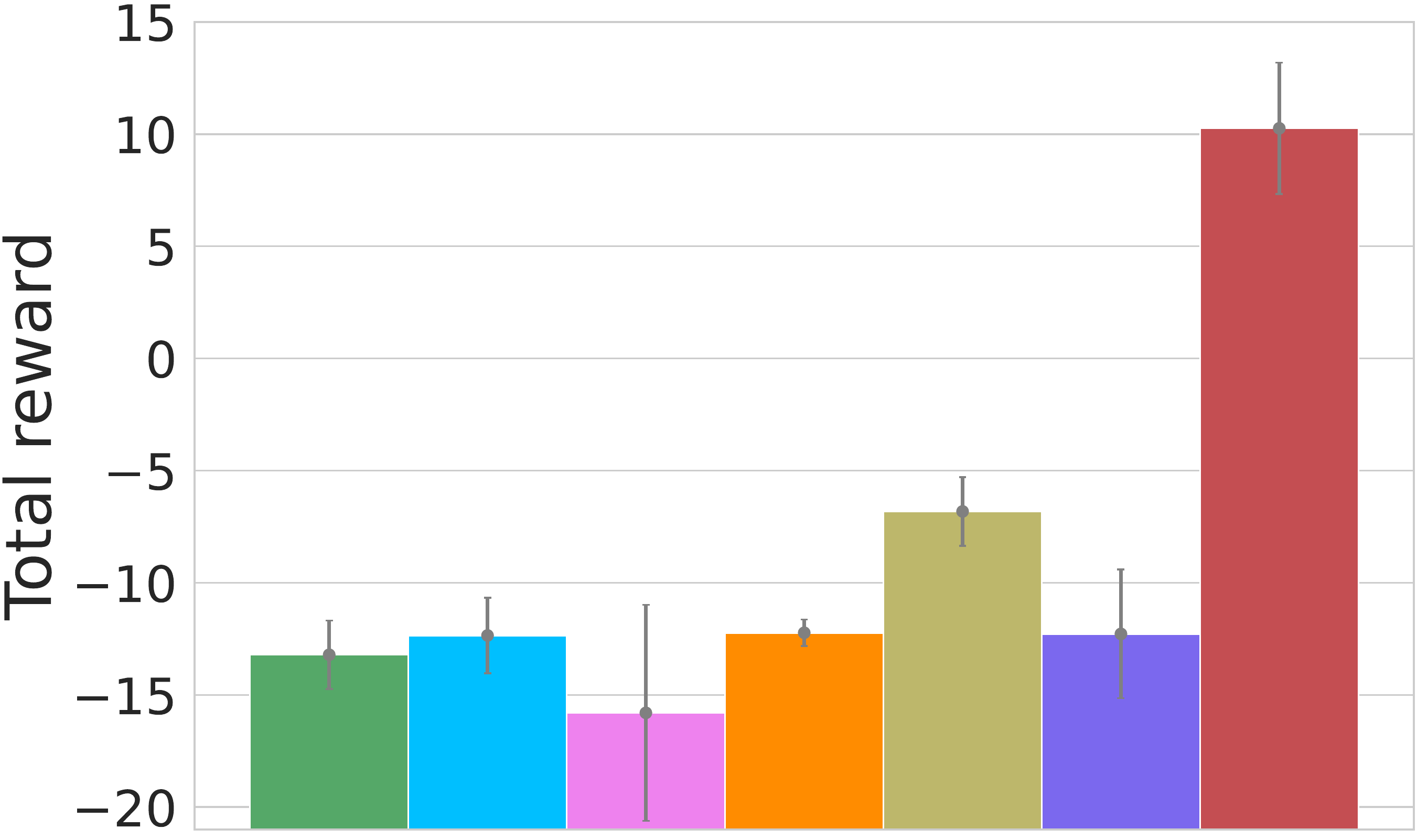}}
    \caption{Comparision of the results between the proposed CSIRL and other IRL baselines on the Mujoco task of \emph{AntUMaze} and the Adroit task of \emph{Relocate}.
    }
    \label{fig::d4rl_result}
\end{figure}

\begin{figure*}[!t]
    \centering
    \subfloat{\quad\includegraphics[scale=0.8]{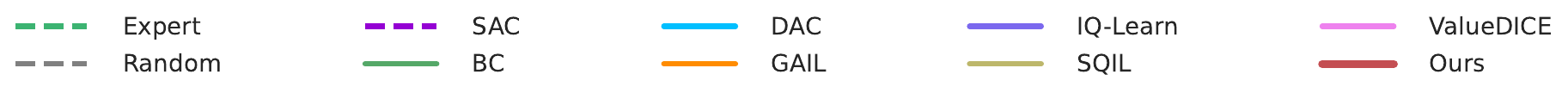}}\\
    \addtocounter{subfigure}{-1}
     \vspace{-0.2cm}
    \subfloat[highway-fast]{\includegraphics[width=0.25\textwidth]{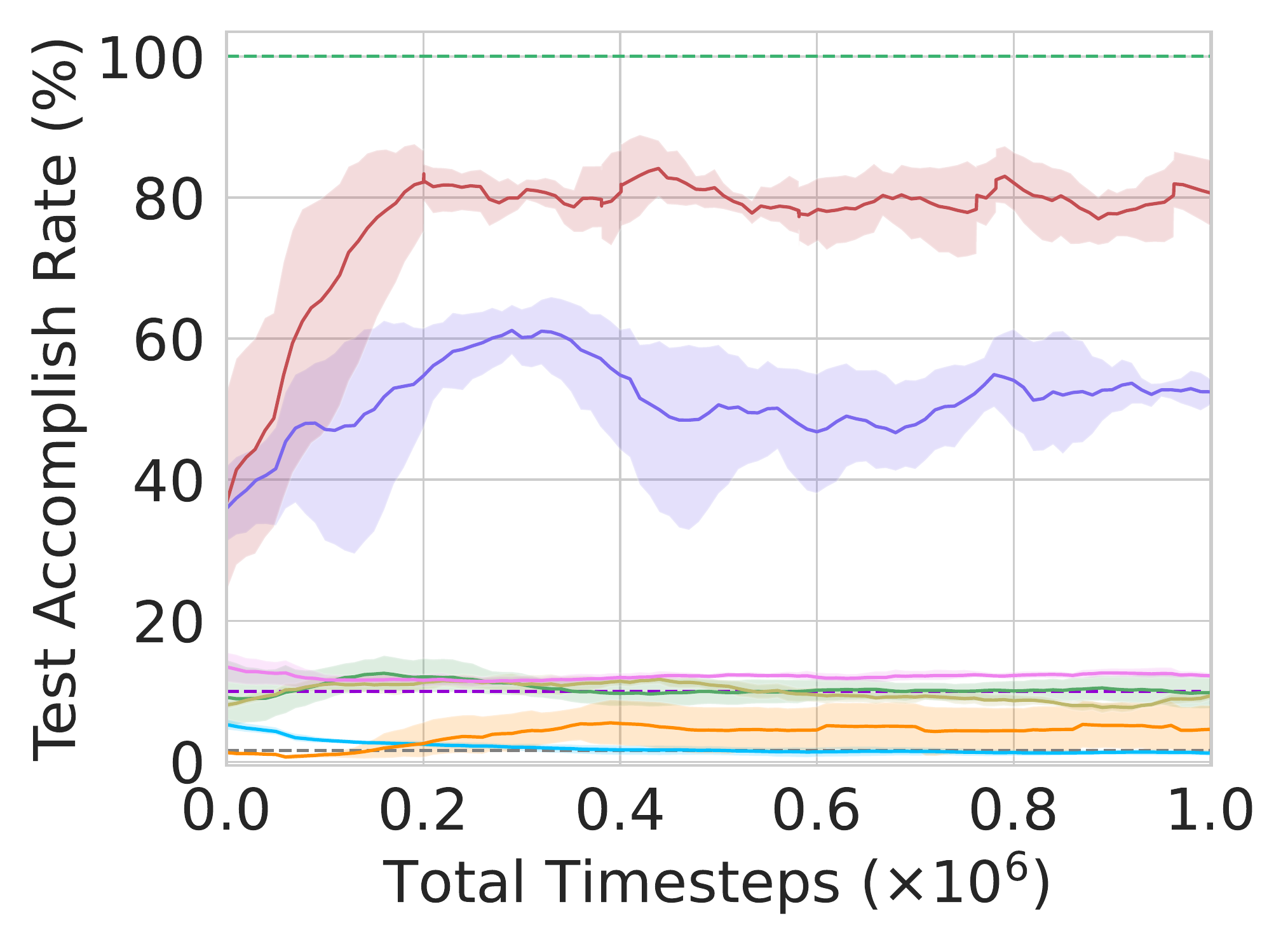}}
    \subfloat[merge]{\includegraphics[width=0.25\textwidth]{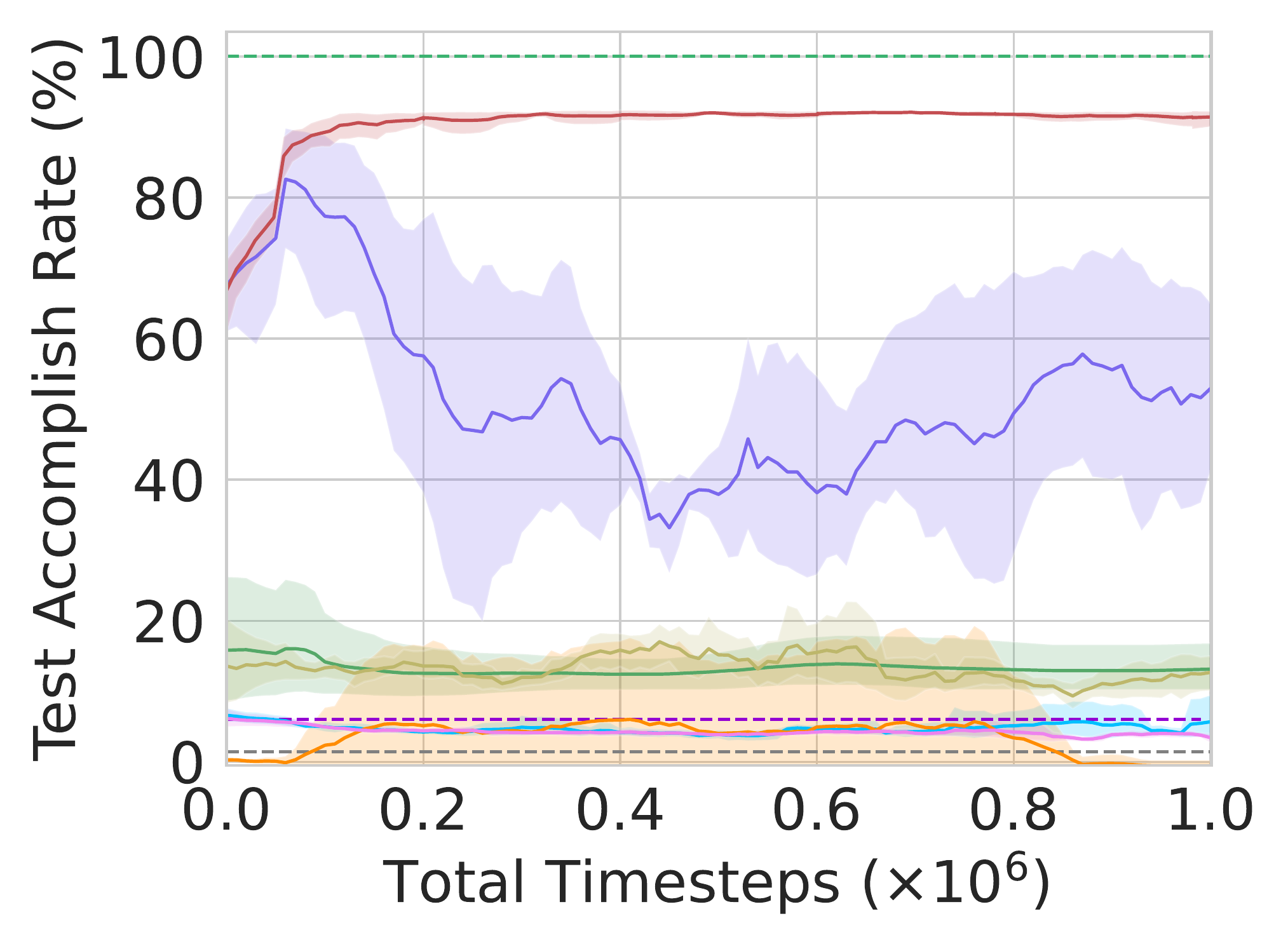}}
    \subfloat[roundabout]{\includegraphics[width=0.25\textwidth]{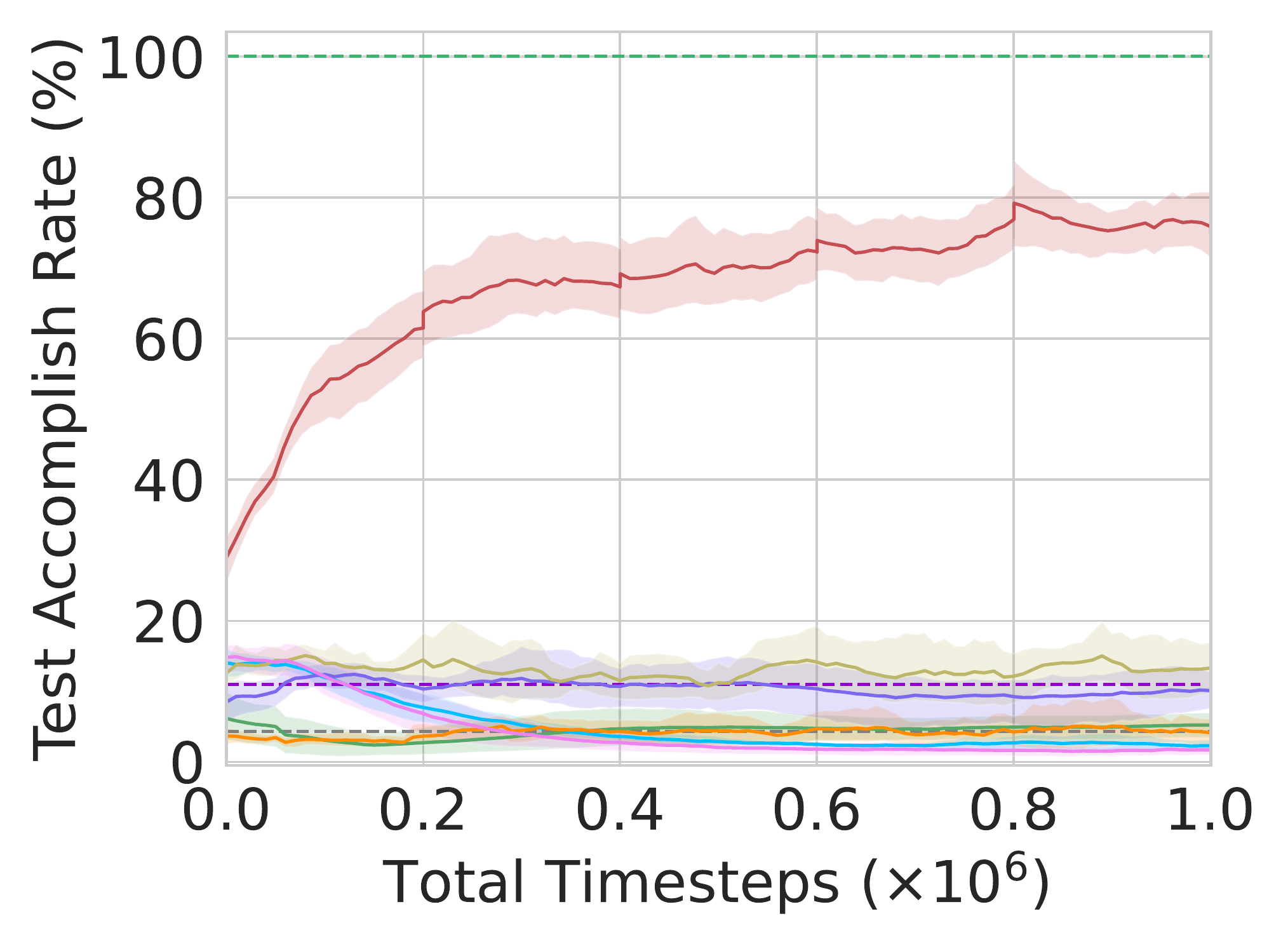}}
    \subfloat[intersection]{\includegraphics[width=0.25\textwidth]{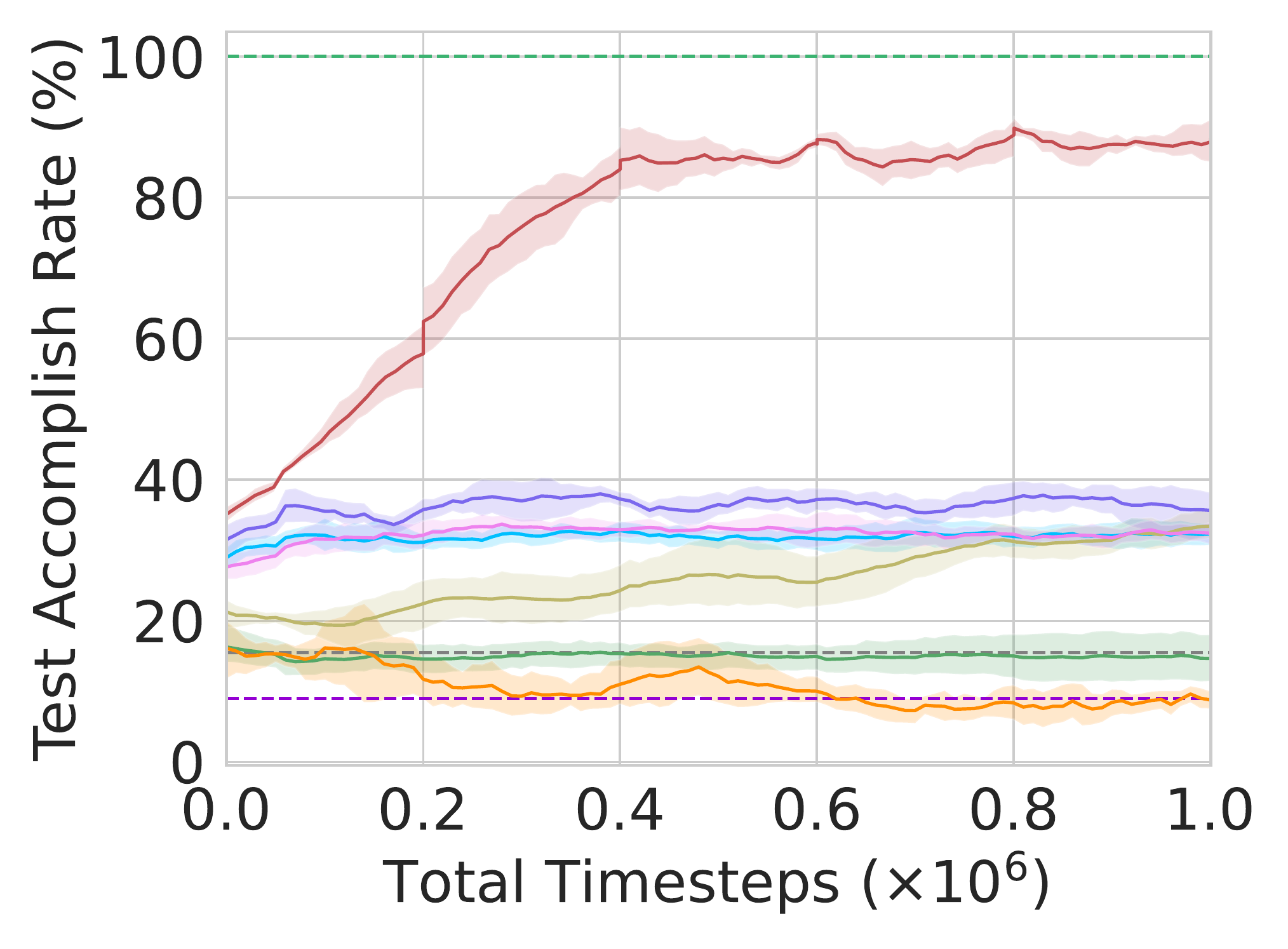}}
    \caption{Learning curves of our proposed CSIRL and baselines. All experimental results are illustrated with the average performance over 5 random seeds for a fair comparison. To make the results clearer for readers, we adopt a 95\% confidence interval to plot the error region.}
    \label{fig::ex1}
  \end{figure*}

  \begin{table*}[!t]
    \vspace{0.3cm}
    \centering
    \caption{Comparison results of our proposed CSIRL and baselines. $\pm$ corresponds to one standard deviation of the average evaluation on 5 random seeds. \textbf{Bold} indicates the best performance in each task. $\dagger$ indicates the performance of non-IRL-based policies. $*$ implies statistical significance for $p < 0.05$ under paired t-test. }
    \resizebox{\textwidth}{!}{%
    \begin{tabular}{@{}clcccc@{}}
    \toprule
     & \multicolumn{1}{c}{\textbf{Method}} & \textbf{\,highway-fast\,}  & \textbf{\,merge\,}    & \textbf{\;intersection\,}          & \textbf{\,roundabout\,}\\ \midrule
    \multirow{10}{*}{\textbf{Test Accomplish Rate}}
    & \textbf{Expert$^\dagger$}  & {1.00 $\pm$ 0.00}  & {1.00 $\pm$ 0.00}  & {1.00 $\pm$ 0.00}  & {1.00 $\pm$ 0.00} \\ \specialrule{0em}{1pt}{1pt}
    & \textbf{Random$^\dagger$}  & {0.02 $\pm$ 0.00}  & {0.01 $\pm$ 0.00}  & {0.16 $\pm$ 0.00}  & {0.04 $\pm$ 0.00} \\ \specialrule{0em}{1pt}{1pt}
    & \textbf{SAC$^\dagger$}~\cite{haarnoja2018soft}  & {0.10 $\pm$ 0.05}  & {0.06 $\pm$ 0.02}  & {0.11 $\pm$ 0.03}  & {0.09 $\pm$ 0.07} \\ \specialrule{0em}{1pt}{1pt}
    \cmidrule(l){2-6} 
    & \textbf{BC}~\cite{pomerleau1991efficient}  & 0.10 $\pm$ 0.03  & 0.13 $\pm$ 0.04  & 0.14 $\pm$ 0.03  & 0.04 $\pm$ 0.02   \\ \specialrule{0em}{1pt}{1pt}
    & \textbf{GAIL}~\cite{ho2016generative}  & 0.06 $\pm$ 0.04  &  0.01 $\pm$ 0.05 & 0.09 $\pm$ 0.02 & 0.04 $\pm$ 0.02\\ \specialrule{0em}{1pt}{1pt}
    & \textbf{DAC}~\cite{kostrikov2018discriminator}  & 0.01 $\pm$ 0.01  &  0.05 $\pm$ 0.02 & 0.32 $\pm$ 0.01 & 0.03 $\pm$ 0.01\\ \specialrule{0em}{1pt}{1pt} 
    & \textbf{ValueDICE}~\cite{kostrikov2019imitation}  & 0.12 $\pm$ 0.03  & 0.04 $\pm$ 0.02  & 0.32 $\pm$ 0.01  & 0.02 $\pm$ 0.01 \\ \specialrule{0em}{1pt}{1pt}
    & \textbf{SQIL}~\cite{reddy2019sqil}  & 0.09 $\pm$ 0.03  & 0.13 $\pm$ 0.02  & 0.32 $\pm$ 0.02  & 0.13 $\pm$ 0.04\\ \specialrule{0em}{1pt}{1pt}
    & \textbf{IQ-Learn}~\cite{garg2021iq}  & 0.52 $\pm$ 0.03  & 0.49 $\pm$ 0.11  & 0.36 $\pm$ 0.02  & 0.10 $\pm$ 0.04 \\ \specialrule{0em}{1pt}{1pt}
    \cmidrule(l){2-6} 
    & \textbf{Ours}  & \textbf{0.85 $\pm$ 0.01$^*$}  & \textbf{0.92 $\pm$ 0.03$^*$}  & \textbf{0.86 $\pm$ 0.01$^*$}  & \textbf{0.74 $\pm$ 0.04$^*$}\\ \specialrule{0em}{1pt}{1pt}
     \midrule
    \multirow{10}{*}{\textbf{Area Under Curve}}
    & \textbf{Expert$^\dagger$}  & {1.00 $\pm$ 0.00}  & {1.00 $\pm$ 0.00}  & {1.00 $\pm$ 0.00}  & {1.00 $\pm$ 0.00} \\ \specialrule{0em}{1pt}{1pt}
    & \textbf{Random$^\dagger$}  & {0.02 $\pm$ 0.00}  & {0.01 $\pm$ 0.00}  & {0.16 $\pm$ 0.00}  & {0.04 $\pm$ 0.00} \\ \specialrule{0em}{1pt}{1pt}
    & \textbf{SAC$^\dagger$}~\cite{haarnoja2018soft}  & {0.10 $\pm$ 0.02}  & {0.06 $\pm$ 0.00}  & {0.11 $\pm$ 0.01}  & {0.09 $\pm$ 0.05} \\ \specialrule{0em}{1pt}{1pt}
    \cmidrule(l){2-6} 
    & \textbf{BC}~\cite{pomerleau1991efficient}  & 0.10 $\pm$ 0.02  & 0.13 $\pm$ 0.03  & 0.15 $\pm$ 0.02  & 0.04 $\pm$ 0.02   \\ \specialrule{0em}{1pt}{1pt}
    & \textbf{GAIL}~\cite{ho2016generative}  & 0.05 $\pm$ 0.03  &  0.03 $\pm$ 0.10 & 0.10 $\pm$ 0.02 & 0.04 $\pm$ 0.02\\ \specialrule{0em}{1pt}{1pt}
    & \textbf{DAC}~\cite{kostrikov2018discriminator}  & 0.01 $\pm$ 0.02  &  0.05 $\pm$ 0.01 & 0.32 $\pm$ 0.01 & 0.05 $\pm$ 0.01\\ \specialrule{0em}{1pt}{1pt} 
    & \textbf{ValueDICE}~\cite{kostrikov2019imitation}  & 0.12 $\pm$ 0.00  & 0.04 $\pm$ 0.03  & 0.32 $\pm$ 0.02  & 0.04 $\pm$ 0.01 \\ \specialrule{0em}{1pt}{1pt}
    & \textbf{SQIL}~\cite{reddy2019sqil}  & 0.10 $\pm$ 0.00  & 0.13 $\pm$ 0.01  & 0.28 $\pm$ 0.03  & 0.13 $\pm$ 0.02\\ \specialrule{0em}{1pt}{1pt}
    & \textbf{IQ-Learn}~\cite{garg2021iq}  & 0.49 $\pm$ 0.05  & 0.51 $\pm$ 0.10  & 0.36 $\pm$ 0.02  & 0.10 $\pm$ 0.03 \\ \specialrule{0em}{1pt}{1pt}
    \cmidrule(l){2-6} 
    & \textbf{Ours}  & \textbf{0.76 $\pm$ 0.02$^*$}  & \textbf{0.90 $\pm$ 0.04$^*$}  & \textbf{0.78 $\pm$ 0.02$^*$}  & \textbf{0.68 $\pm$ 0.03$^*$}\\ \specialrule{0em}{1pt}{1pt} 
    \bottomrule
    \end{tabular}%
    }
    
    \label{tab:result}
\end{table*}
\subsection{Comparison}
For the D4RL benchmark, we present the experimental results of the trained agent from these compared baselines in Figure~\ref{fig::d4rl_result}. Each baseline is provided with 25 expert trajectories. All of the methods are tested with five different random seeds to ensure comparison. In the \emph{AntUMaze} scenario, other IRL baselines like SQIL, DAC and ValueDICE are all not competitive. Meanwhile, IQ-Learn and our CSIRL significantly outperform other IRL baselines by an average factor of 10. Moreover, While IQ-Learn achieves similar performance to our CSIRL in the \emph{AntUMaze} scenario, for the \emph{Relocate} scenario, our CSIRL is the only algorithm that can receive a positive cumulative sum reward. This comparison significantly shows the advantages of our CSIRL over other IRL baselines when facing complex robot controlling problems.

For the autonomous driving benchmark, the experimental results of all compared methods using 25 expert trajectories are shown in Figure~\ref{fig::ex1} and Table~\ref{tab:result}. We adopt the test accomplish rate, the percentage of the driving process towards the final destination, as the evaluation metric. 
In these challenging scenarios, the performance of the random policy is extremely low.
Meanwhile, since the self-driving vehicle has to avoid collision with other on-road vehicles while navigating towards the destination, the naive destination-distance reward is hard for the agent to learn such skills.
In our experiments, we train the backbone algorithm SAC in the DRL field based on the real destination-distance reward. The result that the SAC agent even with the global reward is unable to learn an effective policy supports this conjecture. It further illustrates the randomness and difficulty of the testes tasks.

Several state-of-the-art IRL baselines, including GAIL, ValueDICE, and SQIL, cannot learn any effective policy and only achieve the performances on par with those obtained by the random policy.
Moreover, in the \emph{highway-fast} and \emph{merge} scenarios, IQ-Learn exhibits promising performance at the beginning of training since the task in the early stage is relatively simple. However, as the task progresses, IQ-Learn inevitably falls into the suboptimal policy with a low test accomplish rate.
For the complex \emph{roundabout} and \emph{intersection} scenarios that involve difficult lane changing, IQ-Learn also fails to master the long-term skills and performs poorly.
In contrast, our proposed CSIRL successfully improves learning efficiency and final performance.
The results suggest that CSIRL progressively expands the exploration boundary for agent learning, which helps the agents to reconstruct the local reward functions and achieves non-trivial performance.

\begin{figure}[!t]    
    \centering
    \subfloat{\quad\includegraphics[scale=0.7]{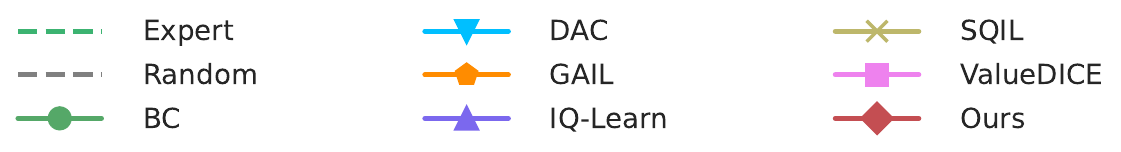}}\\
    \addtocounter{subfigure}{-1}
     \vspace{-0.2cm}
    \subfloat[roundabout]{\includegraphics[width=0.245\textwidth]{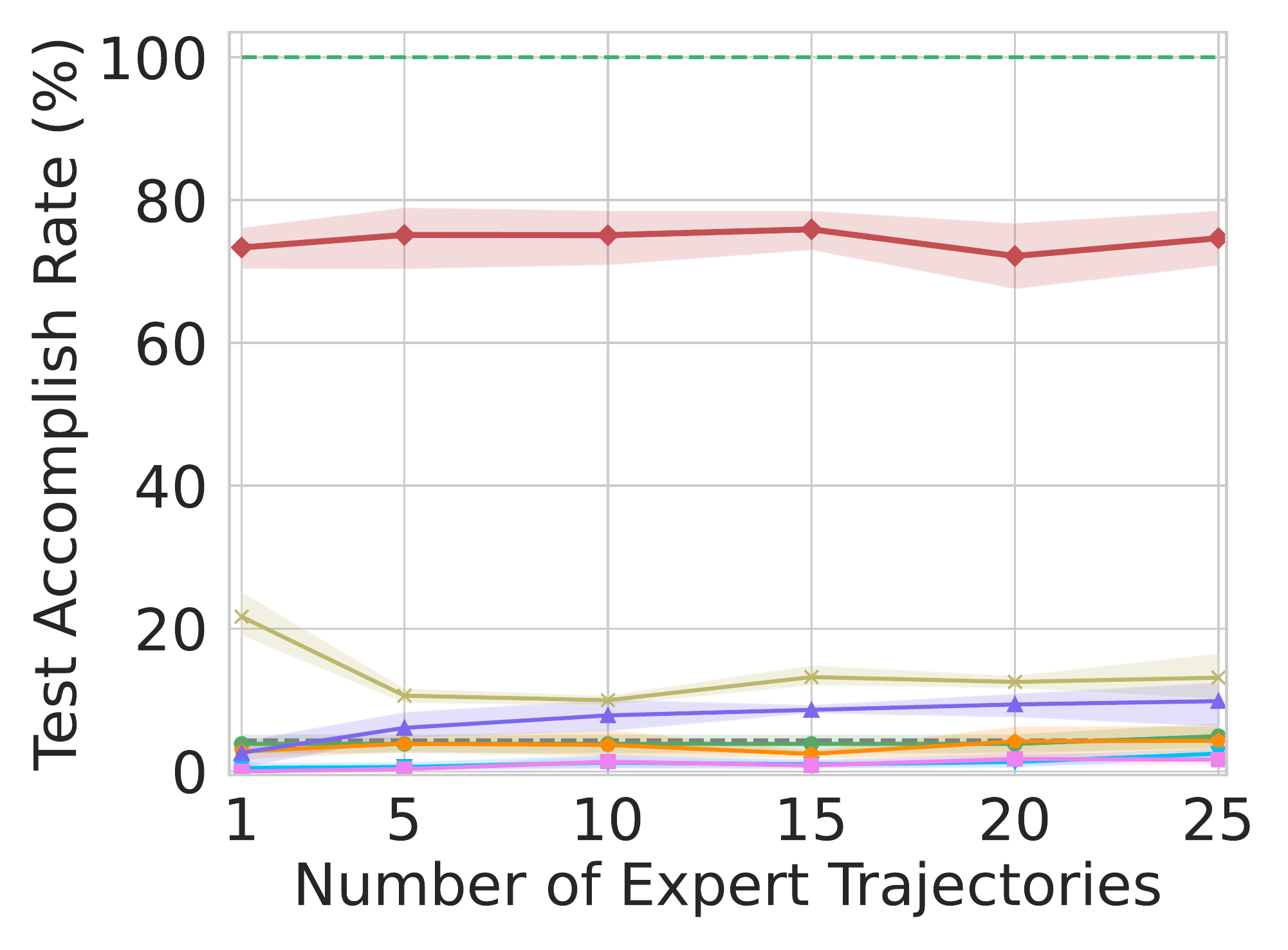}}
    \subfloat[intersection]{\includegraphics[width=0.245\textwidth]{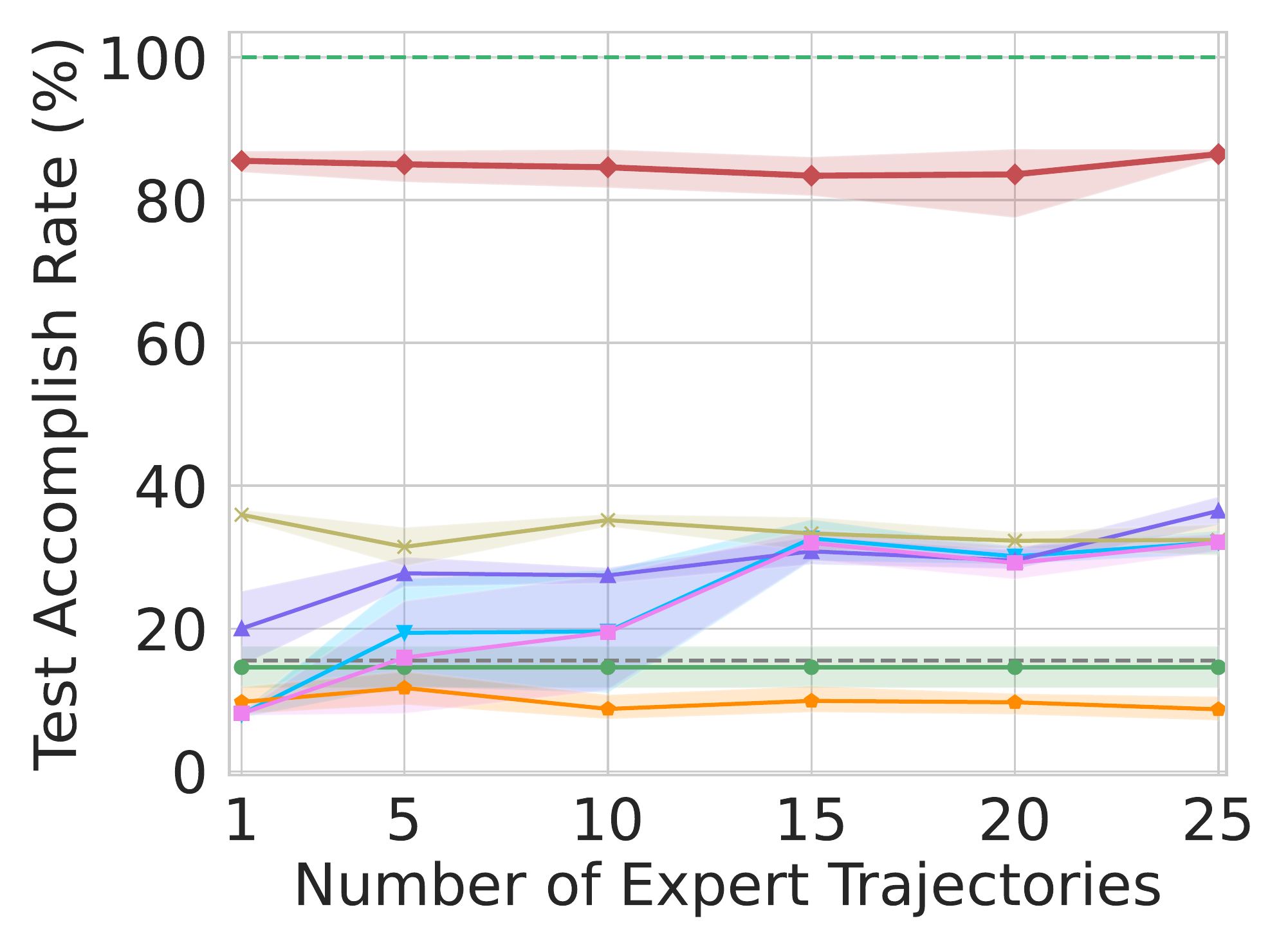}}
    \caption{
    The performance comparison between the proposed CSIRL and baselines with different numbers of expert trajectories.
    }
    \label{fig::ex2}
\end{figure}

We further carry out robustness analysis 
using different numbers of expert trajectories, as shown in Figure~\ref{fig::ex2}.
In the \emph{roundabout} scenario, even with the increasing number of expert trajectories, all compared baselines still perform poorly and only achieve similar performance as the random policy. In the \emph{intersection} scenarios, the performance improvement of the baselines is also not trivial, where the baselines are unable to achieve more than $40\%$ of expert performance. However, CSIRL provides an impressive improvement in the performance over the baselines even with only one expert trajectory, showing its robustness to imitation promoting.

\subsection{Ablation Study}

\begin{figure}[!b]
    \centering
    \includegraphics[width=0.47\textwidth]{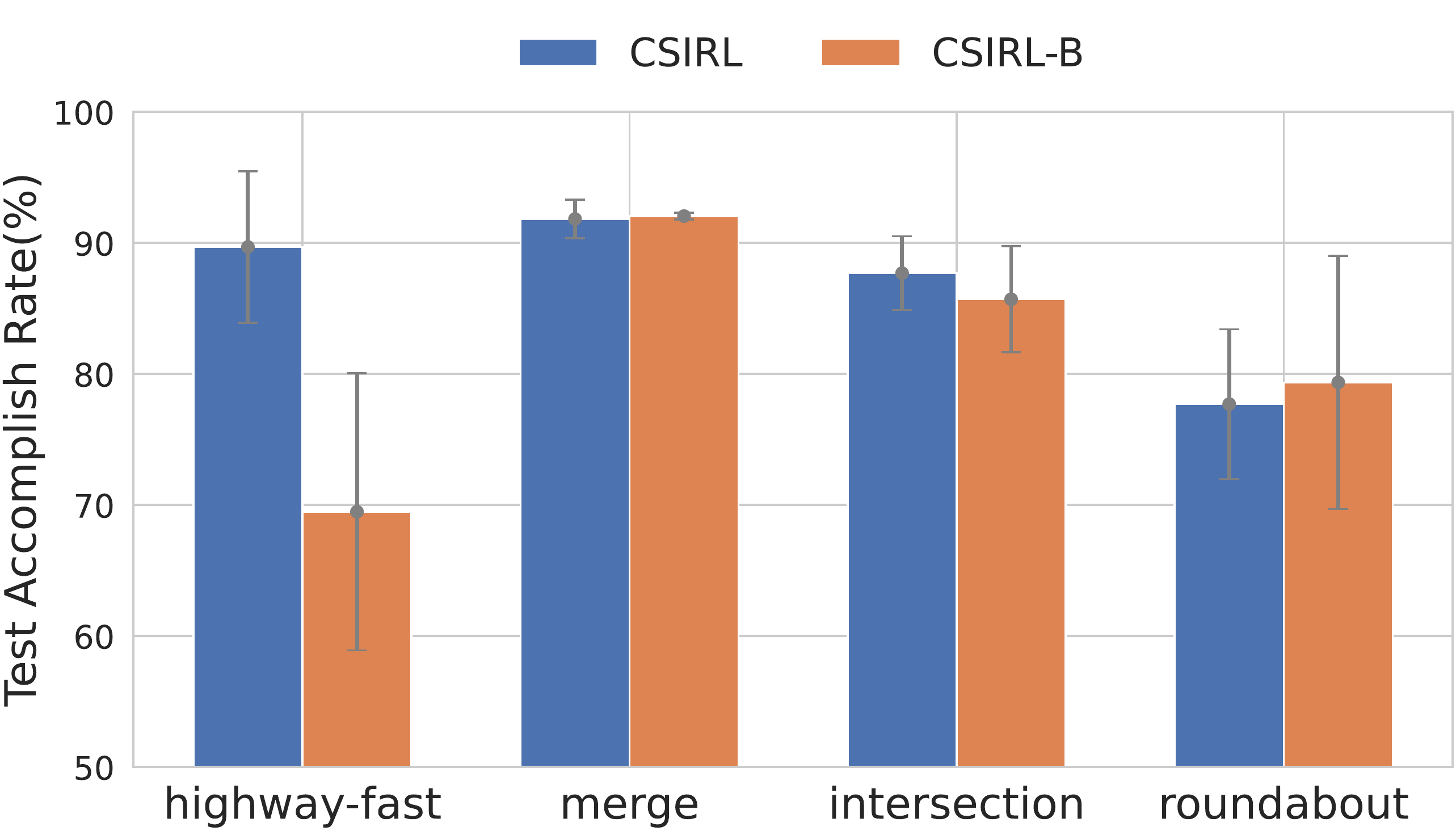}
    \caption{Ablation study on the proposed CSIRL with and without the intrinsic reward generator.}
    \label{fig::ex3}
\end{figure}

To further verify the advantage of the intrinsic reward generator, we implement a variant of CSIRL without the intrinsic reward generator, called CSIRL-B. 
CSIRL-B only performs curricular subgoal selection and directly uses the matching reward to train the agent.
Figure~\ref{fig::ex3} reports the experiment results of CSIRL and CSIRL-B.
In the simple \emph{merge} scenario, CSIRL-B offers a close performance as CSIRL. However, in the more difficult \emph{highway-fast} scenario, CSIRL outperforms the CSIRL-B by a wide margin. Moreover, although CSIRL-B achieves competitive results compared to CSIRL in \emph{intersection} and \emph{roundabout}, CSIRL-B still suffers from a much more significant variance.
The results show that the intrinsic reward generator indeed helps to fine-tune the reward function for expert-like behaviors in long-term goals. By meta-learning, the intrinsic reward generator provides a more diversified reward instead of the matching reward that only considers the subgoal direction. This intrinsic reward reasonably evaluates the states according to expert demonstrations and agent interactions, leading to a better imitation performance.

\begin{figure}[!t]    
    \centering
    \subfloat[roundabout]{\includegraphics[width=0.48\textwidth]{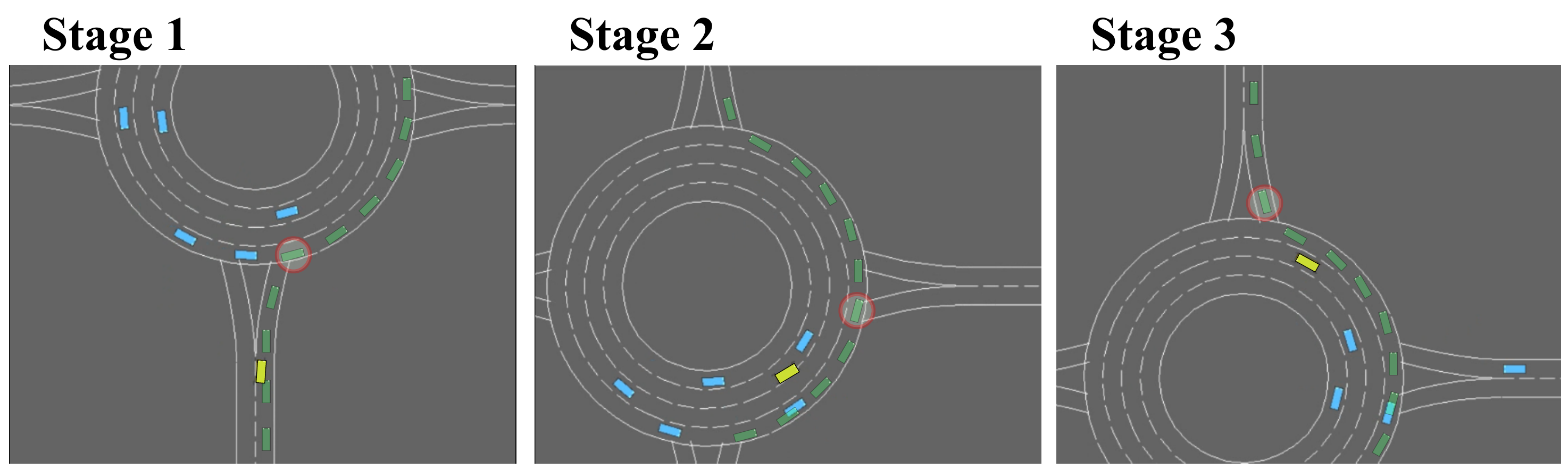}\label{fig::ex4-1}}
    \\
    \subfloat[intersection]{\includegraphics[width=0.48\textwidth]{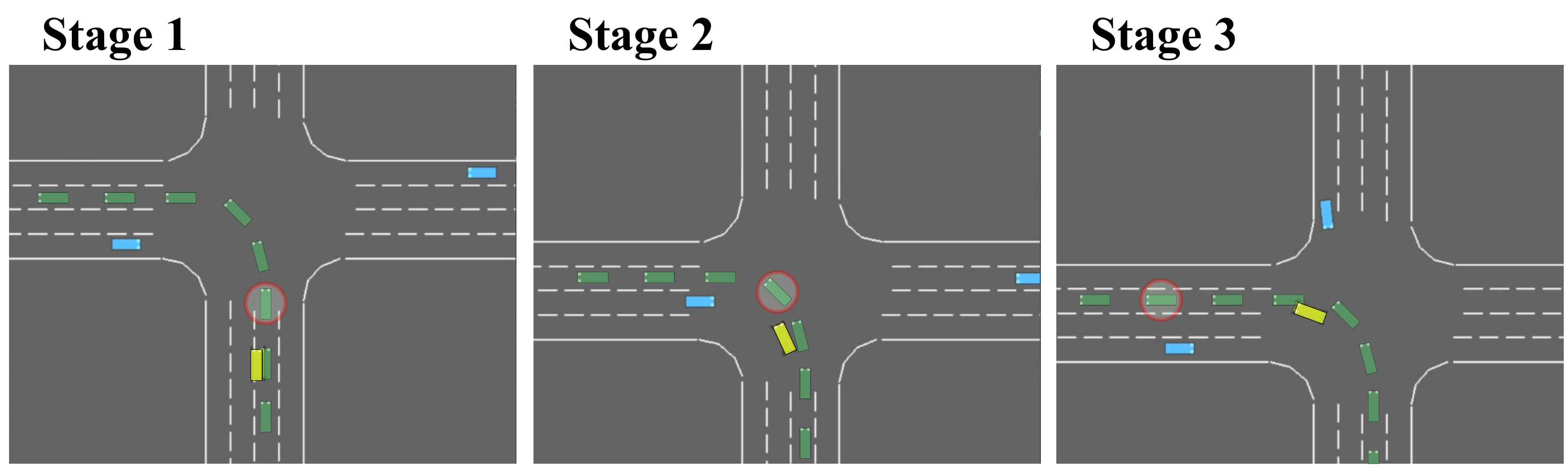}\label{fig::ex4-2}}
    \caption{
    The visualization examples of the selected curricular subgoals in the \emph{roundabout} and \emph{intersection} scenarios. The yellow vehicle is controlled by the agent, while the blue vehicles are controlled by a built-in AI. A series of green vehicles indicate the expert trajectory, where the red circle represents the selected curricular subgoal for the agent.
    }
    \label{fig::ex4}
\end{figure}

\subsection{Subgoal Visualization}
To further explain the subgoal-selecting strategy of CSIRL, we visualize some examples of the curricular subgoals during the training process in \emph{roundabout} and \emph{intersection}, as shown in Figure~\ref{fig::ex4}. 
In the \emph{roundabout} scenario, the agent must learn to go through the roundabout via a specific exit. As shown in Figure~\ref{fig::ex4-1}, the agent first selects the subgoal at the entrance where the controlled vehicle needs to drive into the roundabout. The skill of driving into the roundabout must be learned in this stage. In the next stage, the exploration boundary is expanded, and the subgoal is selected at the first exit of the roundabout, where the agent must keep driving in the roundabout instead of exiting. After that, the subgoal is selected as the second exit of the roundabout, where the agent has to learn to exit the roundabout and go forward to reach the final state.
In the \emph{intersection} scenario, the agent has to drive into the crossroad and exit from it. As visualized in Figure~\ref{fig::ex4-2}, the agent first selects the subgoal at the start position of turning, where the agent has to understand expert makes a turn at the corner instead of going straight ahead. Then the subgoal is selected in the middle position of turning right which requires the skill of taking a sequence of precise and consistent actions. In the next stage, the subgoal is moved to the exiting lane of the crossroad. 
By accomplishing these simple low-level tasks represented by the curricular subgoals, the agent is able to integrate the learned short-term skills to solve the complex long-term task.

\section{Conclusion~\label{sec:conclusion}}
In this work, we propose a novel IRL approach, termed as CSIRL, to disentangle the task progress following a divide-and-conquer scheme. Unlike previous IRL methods that directly search for a global reward function, CSIRL iteratively expands the exploration boundary by selecting curricular subgoals from the expert trajectory to reconstruct local reward functions. Extensive experiments on the autonomous driving benchmark with complex environment dynamics are conducted. The result shows that CSIRL chooses reasonable curricular subgoals as the exploration boundary during training and significantly outperforms the state-of-the-art IRL baselines. In our future work, we will extend our method to learn from a suboptimal expert dataset, which is more challenging in terms of expert policy evaluation.

\textbf{Limitations.} 
Currently, CSIRL focuses on solving multi-stage tasks that can be explicitly decomposed into several subtasks along the task progress. Thus, CSIRL is unsuitable for tasks requiring repetitive short-term skills. It is challenging to extract subgoals from expert demonstrations to represent these tasks. 
Moreover, our method assumes that the expert policy is optimal, otherwise unstable subgoals may conflict with each other, resulting in ineffective exploration. In our future work, we will extend our method to learn from a suboptimal expert dataset, which is more challenging in terms of reliable subgoal evaluation.

\bibliographystyle{plainnat}
\bibliography{reference}

\begin{thebibliography}{59}
\providecommand{\natexlab}[1]{#1}
\providecommand{\url}[1]{\texttt{#1}}
\expandafter\ifx\csname urlstyle\endcsname\relax
  \providecommand{\doi}[1]{doi: #1}\else
  \providecommand{\doi}{doi: \begingroup \urlstyle{rm}\Url}\fi

\bibitem[Andrychowicz et~al.(2017)Andrychowicz, Crow, Ray, Schneider, Fong,
  Welinder, McGrew, Tobin, Abbeel, and Zaremba]{andrychowicz2017hindsight}
Marcin Andrychowicz, Dwight Crow, Alex Ray, Jonas Schneider, Rachel Fong, Peter
  Welinder, Bob McGrew, Josh Tobin, Pieter Abbeel, and Wojciech Zaremba.
\newblock Hindsight experience replay.
\newblock In \emph{{Annual Conference on Neural Information Processing
  Systems}}, 2017.

\bibitem[Aradi(2020)]{aradi2020survey}
Szil{\'a}rd Aradi.
\newblock Survey of deep reinforcement learning for motion planning of
  autonomous vehicles.
\newblock \emph{{IEEE Transactions on Intelligent Transportation Systems}},
  23\penalty0 (2):\penalty0 740--759, 2020.

\bibitem[Bengio et~al.(2009)Bengio, Louradour, Collobert, and
  Weston]{bengio2009curriculum}
Yoshua Bengio, J{\'e}r{\^o}me Louradour, Ronan Collobert, and Jason Weston.
\newblock Curriculum learning.
\newblock In \emph{{International Conference on Machine Learning}}, 2009.

\bibitem[Bhattacharyya et~al.(2022)Bhattacharyya, Wulfe, Phillips, Kuefler,
  Morton, Senanayake, and Kochenderfer]{bhattacharyya2022modeling}
Raunak Bhattacharyya, Blake Wulfe, Derek~J Phillips, Alex Kuefler, Jeremy
  Morton, Ransalu Senanayake, and Mykel~J Kochenderfer.
\newblock Modeling human driving behavior through generative adversarial
  imitation learning.
\newblock \emph{{IEEE Transactions on Intelligent Transportation Systems}},
  2022.

\bibitem[Chan and van~der Schaar(2021)]{chan2021scalable}
Alex~J Chan and Mihaela van~der Schaar.
\newblock Scalable bayesian inverse reinforcement learning.
\newblock \emph{arXiv preprint arXiv:2102.06483}, 2021.

\bibitem[Chane-Sane et~al.(2021)Chane-Sane, Schmid, and Laptev]{chane2021goal}
Elliot Chane-Sane, Cordelia Schmid, and Ivan Laptev.
\newblock Goal-conditioned reinforcement learning with imagined subgoals.
\newblock In \emph{{International Conference on Machine Learning}}, 2021.

\bibitem[De~Haan et~al.(2019)De~Haan, Jayaraman, and Levine]{de2019causal}
Pim De~Haan, Dinesh Jayaraman, and Sergey Levine.
\newblock Causal confusion in imitation learning.
\newblock In \emph{{Annual Conference on Neural Information Processing
  Systems}}, 2019.

\bibitem[Ding et~al.(2019)Ding, Florensa, Abbeel, and Phielipp]{ding2019goal}
Yiming Ding, Carlos Florensa, Pieter Abbeel, and Mariano Phielipp.
\newblock Goal-conditioned imitation learning.
\newblock In \emph{{Annual Conference on Neural Information Processing
  Systems}}, 2019.

\bibitem[Duan et~al.(2017)Duan, Andrychowicz, Stadie, Jonathan~Ho, Schneider,
  Sutskever, Abbeel, and Zaremba]{duan2017one}
Yan Duan, Marcin Andrychowicz, Bradly Stadie, OpenAI Jonathan~Ho, Jonas
  Schneider, Ilya Sutskever, Pieter Abbeel, and Wojciech Zaremba.
\newblock One-shot imitation learning.
\newblock In \emph{{Annual Conference on Neural Information Processing
  Systems}}, 2017.

\bibitem[Eraqi et~al.(2022)Eraqi, Moustafa, and Honer]{eraqi2022dynamic}
Hesham~M Eraqi, Mohamed~N Moustafa, and Jens Honer.
\newblock Dynamic conditional imitation learning for autonomous driving.
\newblock \emph{{IEEE Transactions on Intelligent Transportation Systems}},
  23\penalty0 (12):\penalty0 22988--23001, 2022.

\bibitem[Fang et~al.(2019)Fang, Zhou, Du, Han, and Zhang]{fang2019curriculum}
Meng Fang, Tianyi Zhou, Yali Du, Lei Han, and Zhengyou Zhang.
\newblock Curriculum-guided hindsight experience replay.
\newblock In \emph{{Annual Conference on Neural Information Processing
  Systems}}, 2019.

\bibitem[Florensa et~al.(2018)Florensa, Held, Geng, and
  Abbeel]{florensa2018automatic}
Carlos Florensa, David Held, Xinyang Geng, and Pieter Abbeel.
\newblock Automatic goal generation for reinforcement learning agents.
\newblock In \emph{{International Conference on Machine Learning}}, 2018.

\bibitem[Fu et~al.(2017)Fu, Luo, and Levine]{fu2017learning}
Justin Fu, Katie Luo, and Sergey Levine.
\newblock Learning robust rewards with adversarial inverse reinforcement
  learning.
\newblock \emph{arXiv preprint arXiv:1710.11248}, 2017.

\bibitem[Fu et~al.(2020)Fu, Kumar, Nachum, Tucker, and Levine]{fu2020d4rl}
Justin Fu, Aviral Kumar, Ofir Nachum, George Tucker, and Sergey Levine.
\newblock D4rl: Datasets for deep data-driven reinforcement learning.
\newblock \emph{arXiv preprint arXiv:2004.07219}, 2020.

\bibitem[Garg et~al.(2021)Garg, Chakraborty, Cundy, Song, and
  Ermon]{garg2021iq}
Divyansh Garg, Shuvam Chakraborty, Chris Cundy, Jiaming Song, and Stefano
  Ermon.
\newblock Iq-learn: Inverse soft-q learning for imitation.
\newblock In \emph{{Annual Conference on Neural Information Processing
  Systems}}, 2021.

\bibitem[Haarnoja et~al.(2018)Haarnoja, Zhou, Abbeel, and
  Levine]{haarnoja2018soft}
Tuomas Haarnoja, Aurick Zhou, Pieter Abbeel, and Sergey Levine.
\newblock Soft actor-critic: Off-policy maximum entropy deep reinforcement
  learning with a stochastic actor.
\newblock In \emph{{International Conference on Machine Learning}}, 2018.

\bibitem[Ho and Ermon(2016)]{ho2016generative}
Jonathan Ho and Stefano Ermon.
\newblock Generative adversarial imitation learning.
\newblock In \emph{{Annual Conference on Neural Information Processing
  Systems}}, 2016.

\bibitem[Hren and Munos(2008)]{hren2008optimistic}
Jean-Francois Hren and R{\'e}mi Munos.
\newblock Optimistic planning of deterministic systems.
\newblock In \emph{European Workshop on Reinforcement Learning}, 2008.

\bibitem[Jabri(2021)]{jabri2021robot}
Mohamed~Khalil Jabri.
\newblock Robot manipulation learning using generative adversarial imitation
  learning.
\newblock In \emph{{International Joint Conference on Artificial
  Intelligence}}, 2021.

\bibitem[Jarrett et~al.(2020)Jarrett, Bica, and van~der
  Schaar]{jarrett2020strictly}
Daniel Jarrett, Ioana Bica, and Mihaela van~der Schaar.
\newblock Strictly batch imitation learning by energy-based distribution
  matching.
\newblock In \emph{{Annual Conference on Neural Information Processing
  Systems}}, 2020.

\bibitem[Jeon et~al.(2020)Jeon, Su, Barde, Doan, Nowrouzezahrai, and
  Pineau]{jeon2020regularized}
Wonseok Jeon, Chen-Yang Su, Paul Barde, Thang Doan, Derek Nowrouzezahrai, and
  Joelle Pineau.
\newblock Regularized inverse reinforcement learning.
\newblock In \emph{{International Conference on Learning Representations}},
  2020.

\bibitem[Kaelbling(1993)]{kaelbling1993learning}
Leslie~Pack Kaelbling.
\newblock Learning to achieve goals.
\newblock In \emph{{International Joint Conference on Artificial
  Intelligence}}, 1993.

\bibitem[Kostrikov et~al.(2018)Kostrikov, Agrawal, Dwibedi, Levine, and
  Tompson]{kostrikov2018discriminator}
Ilya Kostrikov, Kumar~Krishna Agrawal, Debidatta Dwibedi, Sergey Levine, and
  Jonathan Tompson.
\newblock Discriminator-actor-critic: Addressing sample inefficiency and reward
  bias in adversarial imitation learning.
\newblock In \emph{{International Conference on Learning Representations}},
  2018.

\bibitem[Kostrikov et~al.(2019)Kostrikov, Nachum, and
  Tompson]{kostrikov2019imitation}
Ilya Kostrikov, Ofir Nachum, and Jonathan Tompson.
\newblock Imitation learning via off-policy distribution matching.
\newblock In \emph{{International Conference on Learning Representations}},
  2019.

\bibitem[Leurent(2018)]{highway_env}
Edouard Leurent.
\newblock An environment for autonomous driving decision-making.
\newblock \url{https://github.com/eleurent/highway-env}, 2018.

\bibitem[Levine et~al.(2011)Levine, Popovic, and Koltun]{levine2011nonlinear}
Sergey Levine, Zoran Popovic, and Vladlen Koltun.
\newblock Nonlinear inverse reinforcement learning with gaussian processes.
\newblock \emph{{Annual Conference on Neural Information Processing Systems}},
  24, 2011.

\bibitem[Lian et~al.(2022)Lian, Kartal, Lewis, Mikulski, Hudas, Wan, and
  Davoudi]{lian2022anomaly}
Bosen Lian, Yusuf Kartal, Frank~L Lewis, Dariusz~G Mikulski, Gregory~R Hudas,
  Yan Wan, and Ali Davoudi.
\newblock Anomaly detection and correction of optimizing autonomous systems
  with inverse reinforcement learning.
\newblock \emph{IEEE Transactions on Cybernetics}, 2022.

\bibitem[Lindner et~al.(2022)Lindner, Krause, and Ramponi]{lindneractive}
David Lindner, Andreas Krause, and Giorgia Ramponi.
\newblock Active exploration for inverse reinforcement learning.
\newblock In \emph{{Annual Conference on Neural Information Processing
  Systems}}, 2022.

\bibitem[Liu et~al.(2022)Liu, Wang, Tian, and Chen]{liu2022learn}
Jinxin Liu, Donglin Wang, Qiangxing Tian, and Zhengyu Chen.
\newblock Learn goal-conditioned policy with intrinsic motivation for deep
  reinforcement learning.
\newblock In \emph{{AAAI Conference on Artificial Intelligence}}, 2022.

\bibitem[Mahmoudieh et~al.(2022)Mahmoudieh, Pathak, and
  Darrell]{mahmoudieh2022zero}
Parsa Mahmoudieh, Deepak Pathak, and Trevor Darrell.
\newblock Zero-shot reward specification via grounded natural language.
\newblock In \emph{{International Conference on Learning Representations}},
  2022.

\bibitem[Markelic et~al.(2011)Markelic, Kjaer-Nielsen, Pauwels, Baunegaard
  With~Jensen, Chumerin, Vidugiriene, Tamosiunaite, Rotter, Van~Hulle, Kruger,
  and Worgotter]{5898415}
Irene Markelic, Anders Kjaer-Nielsen, Karl Pauwels, Lars Baunegaard
  With~Jensen, Nikolay Chumerin, Aušra Vidugiriene, Minija Tamosiunaite,
  Alexander Rotter, Marc Van~Hulle, Norbert Kruger, and Florentin Worgotter.
\newblock The driving school system: Learning basic driving skills from a
  teacher in a real car.
\newblock \emph{{IEEE Transactions on Intelligent Transportation Systems}},
  12\penalty0 (4):\penalty0 1135--1146, 2011.

\bibitem[Mirza and Osindero(2014)]{mirza2014conditional}
Mehdi Mirza and Simon Osindero.
\newblock Conditional generative adversarial nets.
\newblock \emph{arXiv preprint arXiv:1411.1784}, 2014.

\bibitem[Mnih et~al.(2013)Mnih, Kavukcuoglu, Silver, Graves, Antonoglou,
  Wierstra, and Riedmiller]{mnih2013playing}
Volodymyr Mnih, Koray Kavukcuoglu, David Silver, Alex Graves, Ioannis
  Antonoglou, Daan Wierstra, and Martin Riedmiller.
\newblock Playing atari with deep reinforcement learning.
\newblock \emph{arXiv preprint arXiv:1312.5602}, 2013.

\bibitem[Ng et~al.(1999)Ng, Harada, and Russell]{ng1999policy}
Andrew~Y Ng, Daishi Harada, and Stuart Russell.
\newblock Policy invariance under reward transformations: Theory and
  application to reward shaping.
\newblock In \emph{{International Conference on Machine Learning}}, 1999.

\bibitem[Ng et~al.(2000)Ng, Russell, et~al.]{ng2000algorithms}
Andrew~Y Ng, Stuart Russell, et~al.
\newblock Algorithms for inverse reinforcement learning.
\newblock In \emph{{International Conference on Machine Learning}}, 2000.

\bibitem[N{\'u}{\~n}ez-Molina et~al.(2022)N{\'u}{\~n}ez-Molina,
  Fern{\'a}ndez-Olivares, and P{\'e}rez]{nunez2022learning}
Carlos N{\'u}{\~n}ez-Molina, Juan Fern{\'a}ndez-Olivares, and Ra{\'u}l
  P{\'e}rez.
\newblock Learning to select goals in automated planning with deep-q learning.
\newblock \emph{{Expert Systems with Applications}}, 202:\penalty0 117265,
  2022.

\bibitem[Orsini et~al.(2021)Orsini, Raichuk, Hussenot, Vincent, Dadashi,
  Girgin, Geist, Bachem, Pietquin, and Andrychowicz]{orsini2021matters}
Manu Orsini, Anton Raichuk, Leonard Hussenot, Damien Vincent, Robert Dadashi,
  Sertan Girgin, Matthieu Geist, Olivier Bachem, Olivier Pietquin, and Marcin
  Andrychowicz.
\newblock What matters for adversarial imitation learning?
\newblock In \emph{Advances in Neural Information Processing Systems}, 2021.

\bibitem[Paul et~al.(2019)Paul, Vanbaar, and Roy-Chowdhury]{paul2019learning}
Sujoy Paul, Jeroen Vanbaar, and Amit Roy-Chowdhury.
\newblock Learning from trajectories via subgoal discovery.
\newblock In \emph{{Annual Conference on Neural Information Processing
  Systems}}, 2019.

\bibitem[Pomerleau(1991)]{pomerleau1991efficient}
Dean~A Pomerleau.
\newblock Efficient training of artificial neural networks for autonomous
  navigation.
\newblock \emph{Neural Computation}, 3\penalty0 (1):\penalty0 88--97, 1991.

\bibitem[Pong et~al.(2018)Pong, Gu, Dalal, and Levine]{pong2018temporal}
Vitchyr Pong, Shixiang Gu, Murtaza Dalal, and Sergey Levine.
\newblock Temporal difference models: Model-free deep rl for model-based
  control.
\newblock \emph{arXiv preprint arXiv:1802.09081}, 2018.

\bibitem[Puterman(2014)]{puterman2014markov}
Martin~L Puterman.
\newblock \emph{Markov decision processes: discrete stochastic dynamic
  programming}.
\newblock John Wiley \& Sons, 2014.

\bibitem[Racaniere et~al.(2019)Racaniere, Lampinen, Santoro, Reichert, Firoiu,
  and Lillicrap]{racaniere2019automated}
Sebastien Racaniere, Andrew~K Lampinen, Adam Santoro, David~P Reichert, Vlad
  Firoiu, and Timothy~P Lillicrap.
\newblock Automated curricula through setter-solver interactions.
\newblock \emph{arXiv preprint arXiv:1909.12892}, 2019.

\bibitem[Rajaraman et~al.(2020)Rajaraman, Yang, Jiao, and
  Ramchandran]{rajaraman2020toward}
Nived Rajaraman, Lin Yang, Jiantao Jiao, and Kannan Ramchandran.
\newblock Toward the fundamental limits of imitation learning.
\newblock In \emph{{Annual Conference on Neural Information Processing
  Systems}}, 2020.

\bibitem[Rajeswaran et~al.(2018)Rajeswaran, Kumar, Gupta, Vezzani, Schulman,
  Todorov, and Levine]{Rajeswaran-RSS-18}
Aravind Rajeswaran, Vikash Kumar, Abhishek Gupta, Giulia Vezzani, John
  Schulman, Emanuel Todorov, and Sergey Levine.
\newblock {Learning Complex Dexterous Manipulation with Deep Reinforcement
  Learning and Demonstrations}.
\newblock In \emph{{Proceedings of Robotics: Science and Systems}}, 2018.

\bibitem[Reddy et~al.(2019)Reddy, Dragan, and Levine]{reddy2019sqil}
Siddharth Reddy, Anca~D Dragan, and Sergey Levine.
\newblock Sqil: Imitation learning via reinforcement learning with sparse
  rewards.
\newblock \emph{arXiv preprint arXiv:1905.11108}, 2019.

\bibitem[Ross and Bagnell(2010)]{ross2010efficient}
St{\'e}phane Ross and Drew Bagnell.
\newblock Efficient reductions for imitation learning.
\newblock In \emph{{Artificial Intelligence and Statistics}}, 2010.

\bibitem[Ross et~al.(2011)Ross, Gordon, and Bagnell]{ross2011reduction}
St{\'e}phane Ross, Geoffrey Gordon, and Drew Bagnell.
\newblock A reduction of imitation learning and structured prediction to
  no-regret online learning.
\newblock In \emph{{Artificial Intelligence and Statistics}}, 2011.

\bibitem[Schaul et~al.(2015)Schaul, Horgan, Gregor, and
  Silver]{schaul2015universal}
Tom Schaul, Daniel Horgan, Karol Gregor, and David Silver.
\newblock Universal value function approximators.
\newblock In \emph{{International Conference on Machine Learning}}, 2015.

\bibitem[Shahryari and Doshi(2017)]{shahryari2017inverse}
Shervin Shahryari and Prashant Doshi.
\newblock Inverse reinforcement learning under noisy observations.
\newblock \emph{arXiv preprint arXiv:1710.10116}, 2017.

\bibitem[{\v{S}}o{\v{s}}ic et~al.(2018){\v{S}}o{\v{s}}ic, Zoubir, Rueckert,
  Peters, and Koeppl]{vsovsic2018inverse}
Adrian {\v{S}}o{\v{s}}ic, Abdelhak~M Zoubir, Elmar Rueckert, Jan Peters, and
  Heinz Koeppl.
\newblock Inverse reinforcement learning via nonparametric spatio-temporal
  subgoal modeling.
\newblock \emph{Journal of Machine Learning Research}, 19\penalty0
  (69):\penalty0 1--45, 2018.

\bibitem[Sumers et~al.(2021)Sumers, Ho, Hawkins, Narasimhan, and
  Griffiths]{sumers2021learning}
Theodore~R Sumers, Mark~K Ho, Robert~D Hawkins, Karthik Narasimhan, and
  Thomas~L Griffiths.
\newblock Learning rewards from linguistic feedback.
\newblock In \emph{{AAAI Conference on Artificial Intelligence}}, 2021.

\bibitem[Wang et~al.(2021)Wang, Zou, Deng, Tao, Wu, Fan, Chen, and
  Cui]{wang2021reinforcement}
Kai Wang, Zhene Zou, Qilin Deng, Jianrong Tao, Runze Wu, Changjie Fan, Liang
  Chen, and Peng Cui.
\newblock Reinforcement learning with a disentangled universal value function
  for item recommendation.
\newblock In \emph{{AAAI Conference on Artificial Intelligence}}, 2021.

\bibitem[Xu et~al.(2015)Xu, Hu, Jiang, and Meng]{7078931}
Li~Xu, Jie Hu, Hong Jiang, and Wuqiang Meng.
\newblock Establishing style-oriented driver models by imitating human driving
  behaviors.
\newblock \emph{{IEEE Transactions on Intelligent Transportation Systems}},
  16\penalty0 (5):\penalty0 2522--2530, 2015.

\bibitem[Yu et~al.(2019)Yu, Song, and Ermon]{yu2019multi}
Lantao Yu, Jiaming Song, and Stefano Ermon.
\newblock Multi-agent adversarial inverse reinforcement learning.
\newblock In \emph{{International Conference on Machine Learning}}, 2019.

\bibitem[Zakka et~al.(2022)Zakka, Zeng, Florence, Tompson, Bohg, and
  Dwibedi]{zakka2022xirl}
Kevin Zakka, Andy Zeng, Pete Florence, Jonathan Tompson, Jeannette Bohg, and
  Debidatta Dwibedi.
\newblock Xirl: Cross-embodiment inverse reinforcement learning.
\newblock In \emph{{Conference on Robot Learning}}, 2022.

\bibitem[Zheng et~al.(2021)Zheng, Liu, Xu, and Li]{zheng2021objective}
Guanjie Zheng, Hanyang Liu, Kai Xu, and Zhenhui Li.
\newblock Objective-aware traffic simulation via inverse reinforcement
  learning.
\newblock \emph{arXiv preprint arXiv:2105.09560}, 2021.

\bibitem[Zhou and Small(2021)]{zhou2021inverse}
Li~Zhou and Kevin Small.
\newblock Inverse reinforcement learning with natural language goals.
\newblock In \emph{{AAAI Conference on Artificial Intelligence}}, 2021.

\bibitem[Ziebart et~al.(2008)Ziebart, Maas, Bagnell, Dey,
  et~al.]{ziebart2008maximum}
Brian~D Ziebart, Andrew~L Maas, J~Andrew Bagnell, Anind~K Dey, et~al.
\newblock Maximum entropy inverse reinforcement learning.
\newblock In \emph{{AAAI Conference on Artificial Intelligence}}, 2008.

\bibitem[Ziebart et~al.(2010)Ziebart, Bagnell, and Dey]{ziebart2010modeling}
Brian~D Ziebart, J~Andrew Bagnell, and Anind~K Dey.
\newblock Modeling interaction via the principle of maximum causal entropy.
\newblock In \emph{{International Conference on Machine Learning}}, 2010.

\end{thebibliography}

\end{document}